\documentclass[manuscript,sigconf]{acmart}

\AtBeginDocument{%
  \providecommand\BibTeX{{%
    \normalfont B\kern-0.5em{\scshape i\kern-0.25em b}\kern-0.8em\TeX}}}

\setcopyright{acmcopyright}

\copyrightyear{2024}
\acmYear{2024}
\setcopyright{acmlicensed}\acmConference[FAccT '24]{The 2024 ACM Conference on Fairness, Accountability, and Transparency}{June 3--6, 2024}{Rio de Janeiro, Brazil}
\acmBooktitle{The 2024 ACM Conference on Fairness, Accountability, and Transparency (FAccT '24), June 3--6, 2024, Rio de Janeiro, Brazil}
\acmDOI{10.1145/3630106.3659017}
\acmISBN{979-8-4007-0450-5/24/06}

\usepackage{amsmath}
\usepackage{booktabs}
\usepackage{comment}

\usepackage{wrapfig}
\usepackage{fontawesome}

\begin{document}

\title{\textit{Akal Badi ya} Bias: An Exploratory Study of Gender Bias in Hindi Language Technology}

\author{Rishav Hada}
\affiliation{%
  \institution{Microsoft Research}
  \city{Bengaluru}
  \country{India}}
\email{rishavhada@gmail.com}

\author{Safiya Husain}
\affiliation{%
  \institution{Karya}
  \city{Bengaluru}
  \country{India}}
\email{safiya@karya.in}

\author{Varun Gumma}
\affiliation{%
  \institution{Microsoft Research}
  \city{Bengaluru}
  \country{India}}
\email{varun230999@gmail.com}

\author{Harshita Diddee}
\authornote{Work done while at Microsoft.}
\affiliation{%
  \institution{Carnegie Mellon University}
  \city{Pittsburgh}
  \country{USA}}
\email{hdiddee@andrew.cmu.edu}

\author{Aditya Yadavalli}
\affiliation{%
  \institution{Karya}
  \city{Bengaluru}
  \country{India}}
\email{aditya@karya.in}

\author{Agrima Seth}
\authornotemark[1]
\affiliation{%
  \institution{University of Michigan}
  \city{Ann Arbor}
  \country{USA}}
\email{agrima@umich.edu}

\author{Nidhi Kulkarni}
\affiliation{%
  \institution{Karya}
  \city{Bengaluru}
  \country{India}}
\email{nidhi@karya.in}

\author{Ujwal Gadiraju}
\affiliation{%
  \institution{Delft University of Technology}
  \city{Delft}
  \country{Netherlands}}
\email{u.k.gadiraju@tudelft.nl}

\author{Aditya Vashistha}
\affiliation{%
  \institution{Cornell University}
  \city{Ithaca}
  \country{USA}}
\email{adityav@cornell.edu}

\author{Vivek Seshadri}
\affiliation{%
  \institution{Microsoft Research, Karya}
  \city{Bengaluru}
  \country{India}}
\email{visesha@microsoft.com}

\author{Kalika Bali}
\affiliation{%
  \institution{Microsoft Research}
  \city{Bengaluru}
  \country{India}}
\email{kalikab@microsoft.com}

\renewcommand{\shortauthors}{Hada et al.}

\begin{abstract}

Existing research in measuring and mitigating gender bias predominantly centers on English, overlooking the intricate challenges posed by non-English languages and the Global South. This paper presents the first comprehensive study delving into the nuanced landscape of gender bias in Hindi, the third most spoken language globally. Our study employs diverse mining techniques, computational models, field studies and sheds light on the limitations of current methodologies. Given the challenges faced with mining gender biased statements in Hindi using existing methods, we conducted field studies to bootstrap the collection of such sentences. Through field studies involving rural and low-income community women, we uncover diverse perceptions of gender bias, underscoring the necessity for context-specific approaches. This paper advocates for a community-centric research design, amplifying voices often marginalized in previous studies. Our findings not only contribute to the understanding of gender bias in Hindi but also establish a foundation for further exploration of Indic languages. By exploring the intricacies of this understudied context, we call for thoughtful engagement with gender bias, promoting inclusivity and equity in linguistic and cultural contexts beyond the Global North.
\end{abstract}

\ccsdesc[500]{Computing methodologies~Natural language processing}
\ccsdesc[500]{Human-centered computing}

\keywords{Gender bias, Indic languages, Global South, India, Hindi, Community centric}

\received{20 February 2007}
\received[revised]{12 March 2009}
\received[accepted]{5 June 2009}

\maketitle

\section{Introduction}
Large Language Models (LLMs) continue to exhibit increasingly human-like precision across various tasks, leading to their integration into a wide range of real-world applications \cite{song2023restgpt,laskar-etal-2023-building}. 
As these technologies become more readily available and utilized in a multitude of languages, it becomes critical to understand, identify, and address certain critical biases that may appear. Previous research indicates that Natural Language Generation (NLG) models have the potential to generate or intensify biases and this leads to negative impacts on specific user groups and marginalized communities \cite{lucy2023onesizefitsall, trajanovski-etal-2021-text, 10.1145/3449091, 10.1145/3411764.3445372, kirk2021bias}. Gender bias, in particular, is a critical topic of concern. 
Gender biases that exist in language technologies can perpetuate under-representation, stereotyping, or misrepresentation of women and gender minorities \cite{stanczak2021survey}. 
Addressing gender bias in technology is crucial to bridge the digital gender gap and promote a more inclusive and equitable digital society \cite{organisation2018bridging}.

\begin{figure*}[t!]
    \centering
    \includegraphics[scale=0.45]{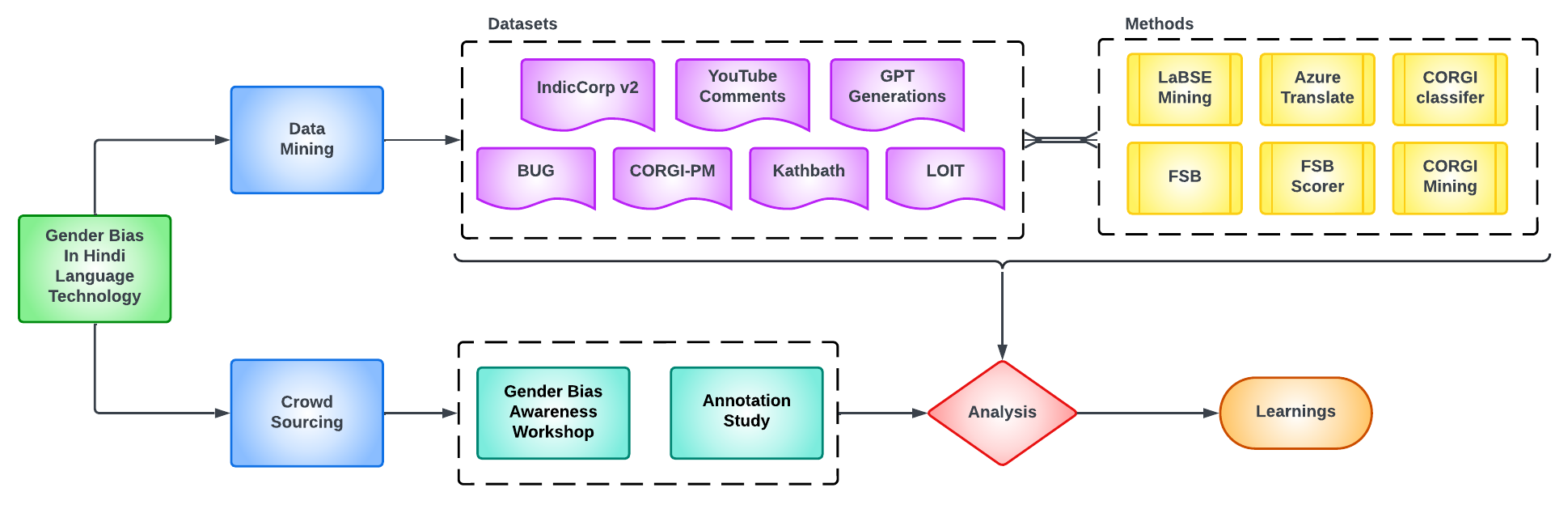}
    \caption{Pipeline of our experiments}
    \label{fig:abb}
\end{figure*}

Due to increasing adoption of language technologies in various languages, it is imperative to understand the biases these models can propagate not only in English but also in different languages and cultures.  
Unfortunately, however, much of the research in measuring and mitigating gender bias is in the context of the English and the Global North. Little is known about how to measure and mitigate gender bias in the context of Global South. The largely understudied dimension of gender bias in the context of Global South specifically for Hindi serves as the primary focus of this work.
Filling this critical gap, we present the first comprehensive study of gender bias in Hindi. Our study highlights particularly in the context of India, it is difficult to utilize the parameters, benchmarks, and guidelines developed for identifying gender bias in English for Indic languages.\footnote{We follow the previous works \cite{gala2023indictrans,doddapaneni-etal-2023-towards} that state Indic languages as a superset, constituting Indo-Aryan, Dravidian, and a few low-resource languages belonging to the Austroasiatic, Sino-Tibetan, and Tai-Kadai families.} Figure \ref{fig:abb} shows the pipeline of our experiments.

We conducted several experiments for mining gender-biased data in Hindi from different sources. Our experiments of mining include lexicon and heuristic-based approaches of mining, computational models for automatic classification of gender bias, and GPT-based generation of biased sentences. We explored data sources like social media comments, news media, and translation of existing gender bias datasets. Our experiments highlight several key challenges in mining gender-biased data in Hindi. We found that a large amount of data available online is Anglo-centric and hence does not serve as a good source for creating gender bias identification dataset in the Indian context. Mining social media data is extremely difficult due to growing restrictions. Heuristic-based approaches return a higher percentage of false positives. Computational models show poor performance due to limited cross-lingual and cross-domain transfer capabilities. Translations from industrial translation systems produce extremely formal and non-contextual translations. Finally, GPT generations show a limited diversity of themes. 

Given the challenges faced with creating a gender bias dataset in Hindi using popular mining techniques proposed in past work for English, we conducted community centered field studies to bootstrap the collection of such sentences. Even though the prevalence of technology is growing in rural India, their opinion is often ignored in development of these technologies. 
For our field studies we employ rural, low income women, to include alternative voices, promoting empowerment within marginalized communities that are disproportionately affected by AI \cite{10.1145/3593013.3594134,10.1145/3551624.3555290,Kormilitzin2023API, 10.1145/3531146.3533132}.
We first aimed to understand what is the shared understanding of gender bias within a Hindi speaking community. The first field study was designed to elicit a culturally relevant definition of gender bias, and crowd-source gender biased statements via activities and plays. In the second field study we conducted a crowd-sourced annotation study to identify varying degrees of gender bias generated by GPT. 
Our first field study revealed variability in perceptions of gender bias. A simple gamified and interactive approach helped in gaining tacit knowledge about gender stereotypes. Our gender bias annotation workshop highlighted the importance of designing annotation tasks while keeping a variety of audiences in mind. The Best-Worst Scaling comparative annotation framework that showed promising results with an urban audience for \citet{hada-etal-2023-fifty} was found to be complex by the rural crowd-workers employed in our study. 

Our findings show that the study of gender bias involves subtle nuances making it a complex topic. Bringing this to an understudied and highly gendered context such as India is even more challenging.
We navigate through the various complexities of identifying gender biased statements in Hindi. The key contributions of our work are as follows:
\begin{itemize}
    \item We conduct in-depth experiments for mining gender biased sentences in Hindi.
    \item We conduct field studies adopting a community centered approach for gender bias identification, and employing rural, low-income community women to foster minority opinion.
    \item We highlight some of the critical challenges faced in mining sentences and field studies that other researchers and technologists should be aware of when engaging in further study of gender bias for Indic languages.
    \item We make recommendations on identifying and mitigating gender bias that focuses on the inclusion of communities right from the beginning. 
\end{itemize}
 We hope that our experiments and case studies can provide a strong foundation upon which to further explore the complex and critical nature of gender bias in Indic languages.\footnote{The title "\textit{Akal Badi ya Bias}" is a word play on the Hindi proverb "\textit{Akal Badi ya Bhains}". The proverb translates to "Is wisdom greater or is the buffalo?" in English. It is used to imply that someone is behaving foolishly or lacking common sense.} \footnote{Code and data available at: \url{https://aka.ms/AkalBadiyaBias}}
\section{Related Work}
    \paragraph {\textbf{Bias Identification:}} Bias Identification is a crucial preliminary step in recognizing their presence in existing models. This can be done by designing special metrics or scoring frameworks to assign a ``bias" score for sentences, documents, or machine-generated synthetic data~\cite{ntoutsi2020bias}. \citet{kaneko-etal-2022-gender} propose a Multilingual Bias Evaluation score, to evaluate bias using English attribute word lists and parallel corpora without requiring manually annotated data. \citet{hada-etal-2023-fifty} generate a dataset of GPT-generated sentences with normative ratings for gender bias and show that bias occurs on a spectrum. Benchmarks such as CrowS-Pairs \cite{nangia-etal-2020-crows} and StereoSet \cite{nadeem-etal-2021-stereoset} aid in measuring various forms of social and stereotypical biases in language models. \citet{zhao-etal-2018-gender,rudinger-etal-2018-gender} release coreference resolution style WinoBias and WinoGender benchmarks and methods to help identify gender bias in existing co-reference resolution systems.  \cite{stanovsky-etal-2019-evaluating} combine the two aforementioned benchmarks and devise an automatic gender bias evaluation method for eight languages with grammatical gender, based on morphological analysis. Similarly, DisCo \cite{50755} is a metric to identify gendered correlations in publicly available pre-trained models. \citet{dev-etal-2022-measures} present a practical framework of harms and a series of questions that practitioners can answer to guide the development of bias measures. \citet{ramesh-etal-2021-evaluating} evaluate gender bias in Hindi-English Machine Translation. 
    They evaluate Google Translate and the Hi-En OpenNMT model for gender bias using existing metrics WEAT and TGBI. 
    \paragraph {\textbf{Debiasing Methods:}} Once the existence of a certain bias is identified in a language model (LM), it is important to address and mitigate the bias  before safely deploying the model in the real world. 
    \citet{kaneko-etal-2022-debiasing} survey different debiasing methods and conclude that extrinsic evaluations, i.e., evaluations that are dependent on LMs performance on a specific task, and intrinsic evaluation measures do not have a strong correlation. 
    \citet{lauscher-etal-2021-sustainable-modular} introduce ``adapter" modules into the original LMs and train the LM on a counterfactually augmented corpus while keeping the rest of the parameters frozen. The authors evaluate this method using both intrinsic and extrinsic measures and show that it is effective in mitigating gender bias in LMs. 
    \citet{barikeri-etal-2021-redditbias} introduce a conversational dataset -- REDDITBIAS -- that can be used to debias LMs for dialog tasks. They show that this dataset allows for bias identification and mitigation across four dimensions: gender, race, religion, and queerness. \citet{10.1145/3377713.3377792} propose a 2 step method for debiasing gender biased Hindi words. First step is to learn the  bias space from set of definitive-gendered word pairs and second is  to measure the biasness of a biased word. 
    They show that their method is useful in decoupling the debiasing process from the word embedding process. \citet{kirtane-anand-2022-mitigating} invesitgate debiasing methods for Hindi and Marathi. They propose debiasing by using partial projection of vectors. Using partial projections overcomes the issue with linear projection where some word vectors which are gendered by definition were changed. In partial projection instead of zero magnitude along the gender direction, they project a magnitude of constant $\mu$ along with it. They show how their debiasing method works with different techniques such as RIPA and PCA.
    \paragraph {\textbf{Challenges:}} Capturing, measuring, and evaluating all types of biases in language models or crowdsourced data has its challenges. \citet{orgad-belinkov-2022-choose} find that only a few extrinsic metrics are measured in most studies and that datasets and metrics are often coupled. They discuss how their coupling hinders the ability to obtain reliable conclusions, and how one may decouple them. \citet{draws2021checklist} argued that cognitive biases in crowdsourced data can also go unnoticed unless specifically assessed or controlled for. \citet{sharifi2023perspective} show how crowdsourced elicitation can be inherently biased by the open and closed-world perceptions of annotators. \citet{zhou-etal-2021-challenges} show that debiasing a model trained on biased toxic language data is not as effective as simply relabeling the data to remove existing biases. Completely mitigating gender bias from models is a hard task as \citet{dev-etal-2022-measures,doi:10.1126/science.aal4230,spinde-etal-2021-neural-media,10.1145/3340531.3412876,DBLP:conf/nips/BolukbasiCZSK16} show that the societal and cultural prejudices are deeply embedded within the training data due to the presence of bias, and these biases, whether explicit or implicit, can significantly impact the functionality of the NLP systems \cite{DBLP:journals/eswa/RazaGRBD24}.
    We refer the readers to \citet{stanczak2021survey} and \citet{10.1145/3531146.3534627} for detailed surveys of Gender Bias in Natural Language Processing. 
    
    Much of the past work in measuring and mitigating gender bias focuses on English and the context of Global North. There is limited understanding of how to measure and mitigate gender bias in the context of the Global South. The emphasis of past work on understanding and mitigating gender bias in western context has resulted in a gap, as western perspectives may not fully capture the nuances of gender bias in the context and languages of Global South.
    In this paper, we study identification of gender bias specifically for Hindi and Indian context. We conduct several experiments to mine gender biased statements in Hindi based on existing methods proposed for English. Existing methods had several limitations when used for Hindi in the Indian context.
    Due to the challenges faced with mining such sentences we conduct field studies and adopt a community centered approach for gender bias identification to bootstrap the collection of such sentences. 
    When it comes to identifying gender bias in Indic languages, including communities that are fluent in these languages is extremely important \cite{abraham-etal-2020-crowdsourcing}. Moreover, for the subjective task of gender bias identification it is crucial to include a diverse set of annotators as perceptions of bias depends on factors such as lived experiences, background, community, and others \cite{hada-etal-2023-fifty}. There is an increasing adoption of technology in rural India, however, their opinions are often unheard and overlooked while these technologies are created \cite{ASER2022}. It is imperative to foster alternative voices and minority opinions during the development of these technologies \cite{hada-beyond, stochastic-parrots, 10.1145/3593013.3594134}. The inclusion of users in the research process presents substantial potential for mitigating power imbalances and fostering empowerment within marginalized communities that experience disproportionate impacts from AI \cite{10.1145/3593013.3594134,10.1145/3551624.3555290,Kormilitzin2023API, 10.1145/3531146.3533132}.
    In such a case, crowdwork, has emerged as a significant sector in the Indian labour market place \cite{demo_dyn_mtutk, Mechanic29:online}, as an effective method to include local Indic language speaking communities. Additionally, from 2018 to 2022 the portion of household in India with smartphones has doubled from  36\% to 74.8\% \cite{ASER2022}. This proliferation of smartphones in India makes crowdwork platforms more accessible on mobile interfaces~\cite{chopra_karya,joshi-etal-2019-unsung, vashistha_respeak} allowing low-income local language speakers with only basic qualifications to benefit from participating in data annotation work. 
\section{Sourcing Hindi Data for Gender Bias Identification}
To mine gender-biased data specifically for Indian context we tried several existing methods that are used to mine gender-biased sentences in English or other languages. In this section, we describe the datasets we explored, our methods, and an analysis of our findings.
\subsection{Datasets}
\label{sec:datasets}
    \paragraph{\textbf{CORGI-PM}:} CORGI-PM is a Chinese corpus for gender bias probing and mitigation \cite{zhang2023corgipm}. The corpus consists of 32.9K sentences with gender bias labels derived by following an annotation scheme specifically developed for Chinese context. To create the corpus, the authors follow an automatic method of data extraction from a raw corpus that gives them potentially gender-biased sentences. These sentences are then annotated for gender bias. Specifically, to create their corpus they follow a two-step filtering process. In the first step, they build a vocabulary of words that have gendered associations. 
    In the second step, they use the list of gendered keywords to recall sentences from the raw corpus. These sentences are then re-ranked, and a threshold for sentence selection is determined. The selected sentences are annotated for gender bias. 
    The corpus contains 22.5K non-biased sentences and 5.2K biased sentences. 
    
    \paragraph{\textbf{IndicCorp v2}:} IndicCorp v2 is a large-scale collection of monolingual corpora for Indic languages  , containing a total of 20.9 billion tokens across 23 Indian languages and English \cite{doddapaneni-etal-2023-towards}. IndicCorp v2 reflects the contemporary use of Indic languages and covers a wide range of topics, primarily crawled from news articles, magazines, and blog posts. The authors source their data from popular Indian-language news websites, discovering most of their sources through online newspaper directories (e.g., w3newspaper) and automated web searches using hand-picked terms in various languages.
    
    \paragraph{\textbf{Kathbath}:} Kathbath is a read speech corpus \cite{10.1609/aaai.v37i11.26521}. The text used to create the corpus is derived from IndicCorp \cite{kakwani-etal-2020-indicnlpsuite}. IndicCorp is a large collection of monolingual corpora consisting of 12 Indic languages collected from diverse Indic-specific sources. The authors take a subset of IndicCorp (~100K sentences) for each of the 12 languages while limiting the sentence length to 8-15 words and allowing only for alphanumeric characters. 
    In this work, we consider the Hindi transcripts of the created corpus to mine biased sentences.  
    
    \paragraph{\textbf{BUG}:} BUG is a large-scale gender bias dataset for coreference resolution and machine translation \cite{levy-etal-2021-collecting-large}. BUG contains 108K English sentences, sampled semi-automatically from large corpora using lexical-syntactic pattern matching. To create the dataset, initially, a syntactic search is performed to identify sentences with challenging syntactic properties across corpora from three domains Wikipedia, Covid-19 research, and PubMed abstracts. Subsequently, the sentences are filtered to ensure they include at least one entity and a corresponding pronoun. The sentence is marked stereotypical or anti-stereotypical for gender roles. Lastly, a manual assessment of BUG is conducted. The dataset consists of 54K stereotypical sentences, $\approx 30K$ anti-stereotypical sentences, and $\approx 24.5K$ neutral sentences. 
    
    \paragraph{\textbf{YouTube Comments}:} 
    Comments or posts on social media platforms represent the thoughts of individuals and communities \cite{hada-etal-2021-ruddit, pushkar-survey}. . Therefore, to study gender bias in the context of India, we collected comments from YouTube via the YouTube API. First, we curated a list of search queries on topics where gender bias data or polarisation was expected. The list of queries are: \textit{"Deepika Padukone Cleavage Controversy"}, \textit{"Hijab ban controversy"}, \textit{"Meninist' Deepika Bhardwaj Has A Few Questions For Feminists In India"}, \textit{"Swara Bhaskar Marriage Controversy"}, \textit{"Manipur Women Paraded"}, \textit{"Kanika Kapoor’s COVID-19 Flak"}, \textit{"SSR and Rhea Chakraborty"}. We took the first 10 videos that appeared for the search queries and extracted up to 105 comments per video. This gave us a total of 7340 comments.
    
    \paragraph{\textbf{Lot of Indic Tweets (LOIT)}:} LOIT is a dataset of Hindi and Telugu tweets\footnote{\url{http://bpraneeth.com/projects/loit}}. LOIT contains most of the Hindi tweets made between 13th January 2017 and 31st December 2018 and Telugu tweets made between 1st January 2010 and 25th June 2019. Twitter allows users to be their natural selves often leading them to portray the biases that they may have than otherwise. 
    
    \paragraph{\textbf{Fifty Shades of Bias (FSB)}:} FSB is a dataset of 1000 GPT-generated English text with normative ratings for gender bias \cite{hada-etal-2023-fifty}. The dataset is created by prompting GPT systematically. The authors first create a seed set of sentences sourced from various corpora. 
    Using the seed set, the authors prompt GPT-3.5-Turbo to either convert or complete a sentence to its gender-biased variation. The generations were then annotated in a comparative annotation setup to assign a gender bias score to each sentence.

\subsection{Experiments}
    \paragraph{ \textbf{LaBSE Mining}:} Inspired by \citet{albalak2022addressing}, we employed an open-retrieval for emergent data collection.
    To analyze the abundance of biased sentences in existing large-scale corpora, we choose the Hindi subset of IndicCorp v2 \cite{doddapaneni-etal-2023-towards} and sample 10M sentences from it. We mine biased sentences from this sample using the LaBSE model\footnote{\url{https://huggingface.co/sentence-transformers/LaBSE}} \cite{feng-etal-2022-language}, and the top 100 biased sentences from FSB as the queries. LaBSE is a language-agnostic embedding model based on BERT \cite{devlin-etal-2019-bert}, which is trained with a contrastive loss and generates close embeddings for sentences that are similar across 109 languages \cite{feng-etal-2022-language}. In our experiment, we first cache the embeddings of all the 10M sentences from the sample using FAISS-DB\footnote{\url{https://github.com/facebookresearch/faiss}}, and query it using the embedding of a source English sentence from FSB. The top 5 most similar sentences (in terms of cosine similarity) for each query in Hindi for a given source were collected, leading to a total of 500 \textit{potentially biased} sentences. From these 500 sentences, we randomly sample 200 sentences (2 per source sentence) and 2 authors of this paper go through 100 sentences each to classify them as "biased" or "not biased". We found that $\approx$ 20\% of the sentences were biased.
    
    \paragraph{\textbf{Translating to Hindi}:} We took a random sample of 100 sentences each from CORGI-PM \cite{zhang2023corgipm} and BUG \cite{levy-etal-2021-collecting-large} datasets. The sample was taken from sentences that were marked as biased. These sentences were translated into Hindi using Azure Translate. One of the authors of this paper annotated the 200 translated sentences to check if the sentence maintains gender bias upon translation and is relevant to the Indian context. We found that 24\% of the translated BUG sentences and 33\% of the translated CORGI-PM sentences remained gender biased in Hindi, and could be used in the Indian context.
    
    \paragraph{\textbf{CORGI classifier}:} We train a binary classifier on the CORGI-PM \cite{zhang2023corgipm} dataset to classify sentences as "biased" or "not biased". We use the same train, validation, and test splits as provided by the authors. We fine-tune mBERT\footnote{\url{https://huggingface.co/bert-base-multilingual-cased}} \cite{devlin-etal-2019-bert} and the corresponding hyperparameters are provided in Table \ref{tab:corgi_classifier_hparams}. The model achieved an accuracy of 0.81 on the test set. We used this binary classifier on the 7340 YouTube comments we extracted. Out of 7340 comments only $\approx 4.67\%  (343)$ comments were classified as biased. We took a random sample of 100 comments from the 343 comments and one of the authors of this paper classified them as "biased" or "not biased". We found that 36\% of the 100 comments were biased.

    \begin{table}[t!]
    \centering 
    \begin{tabular}{@{}ccccc@{}}
    \toprule
    \textit{epochs} & \textit{batch\_size} & \textit{learning\_rate} & \textit{optimizer} & \textit{patience} \\ \midrule
    4 & 128 & 0.00001 & AdamW & 3 \\ \bottomrule
    \end{tabular}
    \caption{Hyperparameters for training CORGI classifier}
    \label{tab:corgi_classifier_hparams}
    \end{table}


    \paragraph{\textbf{FSB scorer}:} Using the 1000 English sentences from FSB \cite{hada-etal-2023-fifty} as a training set, we finetune the IndicBERT v2\footnote{\url{https://huggingface.co/ai4bharat/IndicBERTv2-MLM-Sam-TLM}} model \cite{doddapaneni-etal-2023-towards} using LoRA \cite{hu2022lora}, to avoid over-fitting. IndicBERT v2 is the SOTA NLU model for Indic language and cross-lingual transfer to English. A regression head was appended to the model and the scores predicted were squeezed between -1 and 1, using a \textit{tanh} activation, similar to FSB. Sweeps were conducted for optimal hyperparameters, and Table \ref{tab:fsb_scorer_hparams} provides the final hyperparameters chosen for training. 
    During the final finetuning phase, the model and best hyperparameter configuration achieved an MSE of 0.057 and Pearson correlation of 0.85 on the validation set which is comparable with the score obtained by \citet{hada-etal-2023-fifty}. Using this gender bias score prediction model we obtain a gender bias score for the 7340 comments extracted from YouTube. We sample the top 100 gender bias-scored comments and one of the authors of this paper classified them as "biased" or "not biased". We found that  21\% of the 100 comments were biased.

    \begin{table*}[t!]
    \small
    \begin{tabular}{@{}cccccccccc@{}}
    \toprule
    \textit{epochs} & \textit{lr} & \textit{scheduler} & \textit{optimizer} & \textit{warmup\_steps} & \textit{batch\_size} & \textit{lora\_r} & \textit{lora\_alpha} & \textit{lora\_dropout} & \textit{lora\_modules} \\ \midrule
    30 & 0.001 & linear & AdamW & 40 & 16 & 16 & 32 & 0.05 & {[}"key", "value"{]} \\ \bottomrule
    \end{tabular}
    \caption{Hyperparameters for training the FSB scorer}
    \label{tab:fsb_scorer_hparams}
    \end{table*}

    \paragraph{\textbf{CORGI mining}:} We take a list 684 of Hindi adjectives from the Internet. Using these adjectives, we follow the steps as described by the authors of the CORGI-PM dataset \cite{zhang2023corgipm} to recall sentences from raw corpora. The details are explained in Section \ref{sec:datasets}. From the list of adjectives, we first find adjectives that have a female association. We do this by measuring the dot product of the word embedding of the adjective with the word embedding of 
    ($\vec{Man}$ - $\vec{Woman}$) (we use Hindi words for man and woman). We conduct our experiments with IndicBERT v2 \cite{doddapaneni-etal-2023-towards} and mBERT \cite{Devlin2019BERTPO} to obtain the word representation. 
    For raw corpora, we use the Hindi sentences from Kathbath and a subset of 10M Hindi tweets from the LOIT dataset. Before we filter the sentences, we normalize the scores associated with each Hindi adjective. We set a threshold to select the top female-leaning adjectives and filter sentences from the corpora that contain these adjectives. The mBERT thresholds used to filter 100 sentences each from Kathbath and LOIT are 0.45 and 0.33 respectively. Similarly, the IndicBERT threshold used to filter 100 sentences each from Kathbath and LOIT is 0.904. 
    This gives us a total of 400 sentences.  These filtered sentences are then manually inspected by one of the authors for bias. For sentences retrieved using mBERT we find that 3\% of them were biased from both LOIT and Kathbath. For sentences retrieved using IndicBERT v2 we find that none of them were biased.  
    
    \paragraph{\textbf{GPT Generation}:}  Using the seeds and the prompts provided in FSB \cite{hada-etal-2023-fifty}, we generate 1800 potentially gender biased sentences in English. Six of the authors of this paper went through 300 sentences each and marked if these sentences are "non-sensical" especially in the Indian context and should be removed. We found that $\approx 8\%$ of these sentences were marked as "non-sensical", and the rest could be used in the Indian context, and showed a gradation of gender bias\footnote{We do not annotate these sentences as biased or unbiased because \citet{hada-etal-2023-fifty} show in their work that the generations in FSB have a gradation of gender bias. Instead, upon initial examination, we found that some of the generated sentences did not make sense.}.

\subsection{Results and Analysis}

With the above datasets and methods, we saw varying yields of potentially gender-biased statements in the Indian context. Table \ref{tab:summ_stats} gives a summary of the yield for the different methods and datasets. In this section, we discuss the key challenges faced in using each of these methods and datasets: 

\begin{table*}[t!]
\small
\centering
\begin{tabular}{@{}llcc@{}}
\toprule
\textbf{Dataset} & \textbf{Method} & \textbf{\# of  samples annotated} & \textbf{Yield (in percent)} \\ \midrule
IndicCorp v2     & LaBSE mining     & 200  & 20 \\
BUG              & Azure Translate  & 100  & 24 \\
CORGI-PM         & Azure Translate  & 100  & 33 \\
YouTube comments & CORGI classifier & 100  & 36 \\
YouTube comments & FSB Scorer       & 100  & 21 \\
GPT generations  & FSB              & 1800 & 92 \\
LOIT \& Kathbath (mBERT) & CORGI mining & 200 & 3 \\
LOIT \& Kathbath (IndicBERT) & CORGI mining & 200 & 0 \\
\bottomrule
\end{tabular}
\caption{Yield of potentially gender-biased text from different data sources and methods.}
\label{tab:summ_stats}
\end{table*}

    \paragraph{\textbf{Internet data is Anglo Centric}:} Gathering data from the internet for natural language processing tasks, especially when focusing on Indian languages and topics such as gender bias, poses a significant hurdle. In the context of NLP applications, data collection often involves web scraping. However, it's essential to acknowledge that the majority of data available on the internet tends to align with dominant viewpoints and consists primarily of content in the English language \cite{stochastic-parrots}. This prevalence of English-centric content on popular internet platforms makes it exceedingly challenging to obtain relevant data for languages other than English, particularly those spoken in regions like the global south. For our experiments, we used IndicCorp v2 \cite{doddapaneni-etal-2023-towards} which was a large-scale effort to collect data in Indian languages. In our experiment of LaBSE mining of gender-biased Hindi statements, using 100 gender-biased statements from FSB \cite{hada-etal-2023-fifty} as source we found only 20\% of the examined statements to be biased. A higher yield from this method was expected because we sampled 10M Hindi sentences from IndicCorp v2 and picked 500 sentences that were most similar to very explicitly biased statements from FSB. The low yield could be attributed to the nature of sentences in IndicCorp v2. As mentioned earlier, the corpus was collected by scraping sentences from online news and media sources where the content might already be sanitized and censored.
    
    \paragraph{\textbf{Translation is a bottleneck}:} When data for a particular task and language is not available, the NLP community has often relied on the impressive performance of translation models to translate task data from a high-resource language to the target language. 
    In our experiments, we translated biased statements from the BUG (an English language) dataset \cite{levy-etal-2021-collecting-large} and CORGI-PM (a Chinese language) dataset \cite{zhang2023corgipm} to Hindi. From the biased statements we examined, we found only $\approx 27\%$ of them to maintain their bias after translation. Out of the unbiased sentences, many sentences did not make any grammatical sense after translation or were not contextually relevant. A known challenge in the use of translation models is that they generate excessively formal renditions, potentially diluting the original colloquial or informal nuances present in the source statements\cite{liu-etal-2023-crossing}. For example, "teacher" was translated to "\textit{adhyapak}", and "wife" was translated to "\textit{grihani}". In colloquial Hindi these words are "\textit{teacher}" and "\textit{bahu}" respectively. Furthermore, the nuanced contextual element in the original statements are often not transpositioned completely during the translation process, leading to potential misinterpretations \cite{zhang-toral-2019-effect,gala2023indictrans,vanmassenhove-etal-2021-machine}. Beyond these challenges, variations in linguistic structures, idiomatic expressions, and cultural nuances pose additional complexities, impeding the seamless transference of gender biases across linguistic boundaries \cite{ramesh-etal-2023-fairness}.
    
\begin{figure*}[t!]
    \centering
    \includegraphics[scale=0.5]{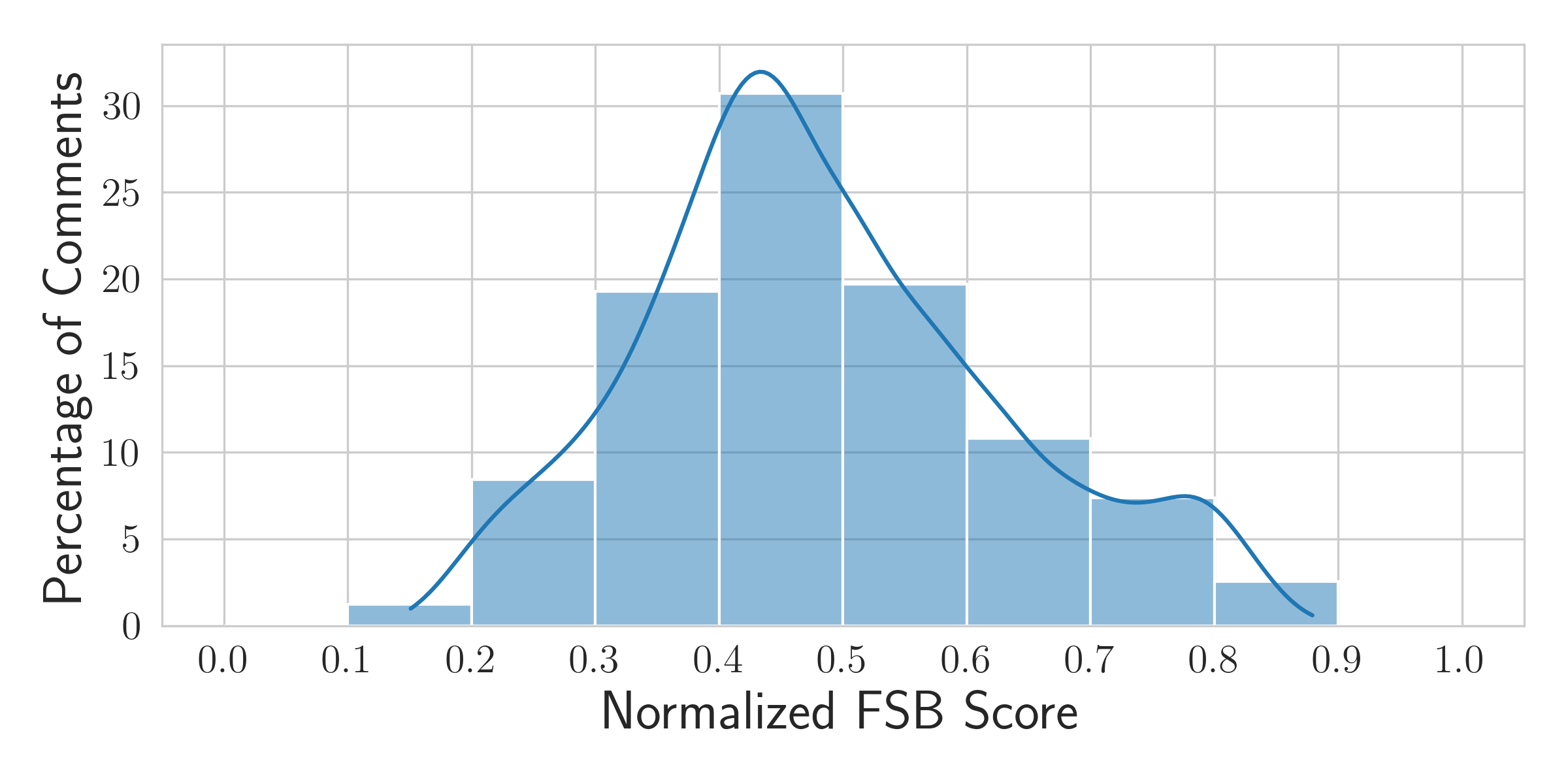}
    \caption{A histogram of the percentage of comments–degree of gender bias. The degree of gender bias scores are grouped in bins of size 0.1.}
    \label{fig:fsb_hist}
\end{figure*}

    \paragraph{\textbf{Problems with collecting social media data}:}  Researchers have traditionally turned to social media platforms like Twitter, Reddit, and Meta when studying concepts like offensive language, hate speech, and identity attacks \cite{hada-etal-2021-ruddit,pushkar-survey, Founta, waseem-hovy-2016-hateful}. Social media data has several advantages like a diverse user base, and users expressing them freely and spontaneously . 
    However, the majority of this data is still in English. For instance, a survey conducted in June 2023 by Statista indicated that approximately 50\% of the desktop traffic on Reddit originates from the United States \cite{reddit-demo}. This prevalence of English-centric content on popular internet platforms makes it exceedingly challenging to obtain relevant data for languages other than English, particularly those spoken in regions like the global south \cite{joshi-etal-2020-state, ahuja-etal-2023-mega}. Data that originates in India is still majorly in English as social media platforms are mostly used by urban population \cite{india-social,india-eng}. Moreover, recent trends in internet platforms have seen an increase in restrictions on data access, making it even more difficult to access social media data for research purposes \cite{davidson2023platform}. This tightening of data accessibility exacerbates the already challenging task of procuring suitable data for gender bias. For our experiments, we extracted 7340 comments from YouTube spanning over 10 controversial/polarising search queries. Our experiment with the CORGI classifier shows that only $\approx 4.6\%$ comments were classified as biased by the classifier, and a random sample of 100 comments showed that only 36\% of those comments were correctly classified as biased. Our experiment with FSB scorer shows that comments from YouTube show a distribution of gender bias score, with a skew towards non-biased or neutral comments as observed in Figure \ref{fig:fsb_hist}. A sample of top-scoring 100 comments showed that only 21\% of these comments are actually biased. Our analysis of random samples also revealed that the majority of the comments are in English, often target individuals over communities, and are highly profane in some cases. Therefore, collecting gender-biased Indian context data from social media platforms is a challenging task as it depends on various factors like appropriate selection of topics, choosing the right signal to boost the representation of biased comments, finding comments in a language other than English, and more. 

    \paragraph{\textbf{Loss due to limitations of cross-lingual or cross-domain transfer of embeddings}:} Recent works \cite{philippy-etal-2023-towards,artetxe-etal-2020-cross,lin-etal-2019-choosing,mhamdi-etal-2023-cross,gala2023indictrans,doddapaneni-etal-2023-towards} have shown the challenges and generalizability of cross-lingual transfer in multilingual language models. We hypothesize that this could be one of the reasons for the poor performance and yield of the CORGI classifier and FSB scorer. CORGI classifier was finetuned and evaluated on Chinese and Hindi respectively.
    Another reason we attribute to their poor performance is the domain mismatch between the training and test data. CORGI was trained with non-social media data but was used to predict scores for YouTube comments. Similarly, the FSB scorer is finetuned with the limited FSB data, which are rudimentary and artificial generations in English created using GPT but is evaluated on the domain of social media content. 
    
    \paragraph{\textbf{Heuristic-based methods of retrieval are challenging}:} 
    In the study of gender bias, researchers have used heuristics such as retrieval from raw corpora using gendered adjectives \cite{zhang2023corgipm}, using professional words \cite{levy-etal-2021-collecting-large}, and template-based sentence creation \cite{nadeem-etal-2021-stereoset,stanczak2021survey}. Our experiment of retrieval of gender-biased statements from raw corpora using gendered adjectives returned only 3\% true positive sentences when using mBERT and 0 true positives when using IndicBERT. We also tried topic-based retrieval from YouTube, coupled with classification using computational models as discussed previously. Both these methods returned a high number of false positives indicating the limitations and challenges of using heuristic-based approaches for retrieval. Additionally, a known limitation of the LLMs is their lack of cultural awareness. While models like LaBSE, IndicBERT, mBERT, and others might be linguistically aware of multiple languages, they lack cultural and social knowledge of different communities primarily because they are not exposed to topcially diverse data from different communities \cite{choi-etal-2023-llms}

    \paragraph{\textbf{Data diversity}:} One of the main components of tasks like gender bias is to have diverse and representative data. The meaning of a sentence is perceived based on one's identity; expression of gender bias in language can often entail cultural nuances, hence, it is equally important that the data is contextually situated \cite{blodgett-etal-2020-language, davani-etal-2022-dealing, biester-etal-2022-analyzing}. 
    In our experiments, we tried various data sources such as IndicCorp v2 which is made primarily of news articles, comments from socia media sites, and text generation from GPT. Generated sentences from GPT showed the most promise in terms of getting a high proportion of biased statements. However, the generated data has limited diversity. We generate up to 4 sentences per seed provided in FSB. The themes are repetitive, sentences do not capture Indian cultural nuances, and are very simple hence do not have idiomatic or linguistic diversity either. The biggest challenge in generating data with this method is a careful selection of diverse seed sets, as the seed sentences can have a significant impact on the generation of biased sentences.

\section{Crowdsourcing Contextually Relevant Gender Bias Definition and Annotations}
\label{sec:crowd}
Due to the difficulties encountered in developing a dataset on gender bias in Hindi using established mining techniques previously suggested for English, we conducted field studies to bootstrap the compilation of such sentences. We do this by adopting a community-centered approach for gender bias identification. This section explores two field studies we conducted in the pursuit of understanding how to define and identify gender bias in Hindi. 

\subsection{Case Study 1: Gender Bias Awareness Workshop - Lucknow, UP}
\label{sec:cs1}
In this field study we aimed to gain a shared understanding of gender bias within a specific community and crowd-source gender bias sentences in Hindi that could inform our data collection process. Definitions of gender bias cannot be constant across cultures and contexts. Global North definitions of gender bias in particular, should not be used to assess situations and environments in the Global South. \citet{braff2022chapter} highlight how the Global North has played a critical role in shaping gender norms and structures as they are often spoken about today. These are however neither universal nor natural. 
It is therefore crucial to consider the cultural, political, and economic contexts of each region when addressing gender issues. 
In this case study where gender bias needed to be spoken about to women, a certain challenge was met in finding the right words and expressions to accurately convey messaging. As an example, the Hindi word for “bias” in the way it is understood in English does not exist; rather commonly used words for this topic instead indicate “discrimination” “favoritism” or “exclusion”.  To this end, we conducted a qualitative experiment to test whether definitions of gender bias in Indian languages could be generated through community-centered methods. 
Working in collaboration with the Milaan Foundation, a prominent non-governmental organization in India that has been working on women’s empowerment and gender bias for over 20 years, we developed a Gender Bias Awareness Workshop. The workshop was held in Lucknow district of Uttar Pradesh state in Northern India with 14 women aged 18-24. Most of these women were enrolled in their bachelor’s degree programs and had also been a part of previous gender-related activities held by the Milaan Foundation. All workshop facilitation and engagement from participants occurred in Hindi to ensure the capture of language-relevant expressions of gender bias. The workshops were structured around two primary components: an exploration of fundamental concepts and the implementation of four key activities to ensure inputs and involvement from the community. In the conceptual exploration participants delved into the distinctions between sex and gender, analyzed gendered language within the context of Hindi, understood the nuances of gendered and non-gendered explanations, identified prevalent instances of gender bias in the daily lives of Indian women, and shared their personal experiences of gender bias. Complementing these discussions were four interactive activities: a storytelling exercise involving two frogs to uncover backstories, skit and role-play scenarios portraying gender dynamics in various situations, a collaborative compilation of a list highlighting bias towards women, the creation of a shared definition of gender bias. 

\subsection{Case Study 2: Annotation Study - Kannauj, UP}
We conducted an annotation study in the Kannauj district of the Northern Indian state of Uttar Pradesh to investigate the identification of gender bias in sentences generated by GPT-3.5-Turbo. \citet{hada-etal-2023-fifty} show in their work that humans can identify gender bias to varying degrees. They use an efficient comparative annotation framework called Best--Worst Scaling (BWS) to obtain a degree of gender bias score per statement. They argue that BWS has shown to be very effective for annotation of subjective tasks such as offensive language and show it can be extended to gender bias.  We followed the data generation and annotation process described in their work. \citet{hada-etal-2023-fifty} have their data annotated by an urban population, and highlight the importance of incorporating subjective judgments for this task.
In contrast, we have our data annotated by a rural population, to include minority opinion for such tasks. Specifically, the study focused on employing women from rural and low-income communities to annotate a Hindi text corpus and identify sentences with gender bias. The overarching goal was to explore the feasibility of employing low-income women to generate foundational annotated datasets that could support the automation of gender bias detection and mitigation in Hindi corpora. This field study hypothesized that by employing women from Hindi-language speaking communities to highlight bias in GPT-generated corpora, identified sentences would be representative of the biases, stereotypes, and marginalization that the larger population of Hindi-speaking women may face. 
\textbf{Study Structure \& Participants:}
This study framed the annotation of gender bias sentences as digital work tasks that were deployed through the Karya App. 2000 English sentences were generated by GPT-3.5-Turbo using the prompting strategies described in \cite{hada-etal-2023-fifty} to create a corpus of sentences. Specifically, given a seed statement, we prompted GPT-3.5-Turbo to generate a gender-biased completion or conversion of the statement. We randomly sample sentences from COPA as our seed statements. 
These gender-biased generations were then translated into Hindi. We initially used machine translation systems, however these did not yield positive results as the outputs did not pass manual quality checks. The final sentences for the task were manually translated for this study to ensure higher quality. This translated corpora was used to create 2N 4-tuples for the BWS setup as described in other works \cite{hada-etal-2023-fifty, hada-etal-2021-ruddit, bws_jj, kiritchenko-mohammad-2017-best, omre, RePEc:elg:eechap:14820_8, pei-jurgens-2020-quantifying, kiritchenko-mohammad-2016-capturing}. The tuples were randomly assigned to each participant.
The participants were shown a tuple and asked to identify which sentence was “most biased” and which sentence was “least biased”. Each participant was given 261 tuples to annotate. 
Participants of this study involved 15 women from low-income backgrounds who underwent a two-hour virtual training delivered by the research team. All participants were native or fluent Hindi speakers who identified as women, and 46\% were aged 18-25. The majority reported being part of a marginalized caste (93\%), practicing Hinduism (94\%), and being married (74\%). Education levels varied significantly, with 33\% of women not passing the 10th grade and 30\% with a High School Diploma or Bachelor’s degree. Income levels were low, with 66\% coming from households earning less than INR 12,000 (USD 144) per month. Most women reported being unemployed and without a steady income; when working, these women were predominantly engaged in agricultural or domestic work.

\subsection{Results and Analysis}

\paragraph{\textbf{Gender Bias Awareness Workshops}}
During the initial stages of the workshop, participants expressed confidence in understanding gender bias, perceiving no need for discussions on the distinctions between sex and gender. As sessions progressed, however, it became evident that many participants were inadvertently conflating these concepts. Throughout the workshop, participant enthusiasm remained high, with active engagement and occasional challenges to facilitators on the concepts they brought forth. 
The frog exercise worked well in exposing prevalent gender stereotypes as participants assigned stereotypical attributes to the frogs based on gender. For the “female” frog, the participants were more likely to use adjectives such as “afraid”, “weak” and “unsure”; whereas for the “male” frog, words like “brave” and “curious” were assigned. Interestingly for the “male” frog, perceptions were not as constant among the participants. Some participants assigned the male frog adjectives with more negative connotations such as “stupid” and “concerned only about seeming courageous”.
In the skit and role-play activities, a shift towards more equitable portrayals emerged, indicating a loosening of rigid gender norms. About 75\% of the paired teams decided to actively portray scenes that depicted equity and parity between men and women across various scenarios. Skits where “bias of roles” were still present were caveated through a warning that the skit represented the norms and ways their parents or people in the community would act. There became an interesting emphasis at this point on what was “right” and what was a “wrong” way to speak about people, with the overwhelming conclusion being that gender has nothing to do with why a person might behave a certain way.  
The creation of a shared definition posed challenges, primarily due to participant fatigue, yet the exercise yielded valuable insights, uncovering over 80 unique sentences illustrating various gender biases, phrases, and words contributing to women's marginalization in everyday scenarios. Despite the difficulty in crystallizing a definitive definition, the workshop successfully captured nuanced perspectives on gender bias.

\paragraph{\textbf{Annotation Study}}
The annotation study revealed several critical insights surrounding the scalability of this type of work. Key challenges occurred in achieving agreement among annotators, with initial agreement values around 0.08, indicating random annotations by the participants. After filtering out annotations completed in less than 30 seconds, a slight improvement to 0.11 in agreement was observed. However, 30\% of annotations selected the same item for both "most" and "least" biased sentences, suggesting issues with the task structure and understanding.
Interviews with participants confirmed the supposition that participants had difficulty in completing the task. 
Most of the women highlighted that they did not understand the virtual training that occurred. While some were not prepared to be connected via a virtual mode facing challenges connecting through video conferencing, others found the language and overall explanation of the tasks too confusing and difficult to follow. 
Further probing highlighted sentence construction and comprehension as a significant barrier; women either found the formal Hindi used in the sentences difficult to understand or felt that the sentences were “nonsensical”. A full manual review of the dataset by native Hindi speakers found that approximately 30\% of sentences were identified as not aligned with the cultural and social contexts of rural India. Roughly 11\% of the sentences did not have a clear meaning when translated, examples of these include “the country discovered new land”. 17\% included words like “dating” or “cowboy”, which are not commonly found in the Indian vocabulary. A precipitating point of confusion was the formality of translation, many of the words in this grouping were transliterated in the annotated corpora which led to misunderstanding amongst the annotators. A few sentences (\~2\%) described scenes or turns of phrase that were unfamiliar to the Indian context in general, such as “thanksgiving”, “bachelorette” and “mowing the lawn”. 
Results from this field study indicate that critical revisions in design and process are required to make this type of task understandable and accessible to participants. The key areas identified for improvement are three-fold. First, establishing more intensive training elements and administering pre-testing to ensure task completion at quality. Second, ensuring that GPT-generated sentences match the social and cultural contexts of the Hindi-speaking community. Third, creating a simpler and more intuitive task structure and expectations. 
\section{Discussion} 
Bring all findings together, our work suggests that studying gender bias especially in the context of the global south or more specifically India can be extremely challenging. Our extensive efforts, spanning experiments and on-field studies, showcase the intricate landscape of gender bias, revealing the need for nuanced strategies and inclusive approaches. We tried several methods of mining gender-biased data from different sources. Our experiments with mining gender-biased sentences highlighted various difficulties. Internet data being very Anglo-centric does not serve as a good source for mining gender biased data for the global south. Extracting data from social media has become increasingly difficult due to growing restrictions. The data we extracted from YouTube based on certain topics had comments majorly in English. From the random sample we annotated based on results from classifiers or gender-biased scoring, only a small portion was actually biased indicating the need for a more careful selection of topics, and domain-specific computational models. We used IndicCorp v2 to mine gender-biased sentences in Hindi. Our experiments with LaBSE embedding and explicitly gender-biased statements, a common method to retrieve similar sentences in two different languages, returned many false positives. This could be attributed to the fact that IndicCorp v2 has been sourced primarily from news outlets containing sentences that are sanitized and censored. Our experiments also highlighted the limitations of current language embeddings, showing limited performance in cross-lingual and cross-domain transfer capabilities. We found using translation systems to be a bottleneck due to their inability to translate sentences contextually and colloquially. We also found keyword-based approaches to retrieval to be ineffective. Finally, the generation of gender biased sentences from GPT showed some promise, however, had a very limited diversity. 

Challenges faced with mining gender biased sentences in Hindi motivated us to conduct field studies to crowd-source such sentences. To incorporate real-world perspectives we delve into the nuanced interpretation of gender bias by conducting two on-field studies. Our case studies bridge the gap between controlled experiments and the dynamic realities of crowd-sourced, culturally relevant insights. We conducted on-field studies to crowd-source gender biased Hindi statements with culturally relevant definition of gender bias and a comparative annotation task for identifying gender-biased statements. The gender bias awareness workshop revealed several key takeaways. The initial overconfidence expressed by participants in understanding gender bias, juxtaposed with their evolving comprehension throughout the workshop, underscores the subtle complexity of the issue. An interactive gamified frog exercise helped us tap into the tacit knowledge of the workshop participants by revealing ingrained gender stereotypes, emphasizing the pervasive nature of biased attributions. The skit and role-play activities provided a platform for a positive shift, with a significant majority actively portraying scenes that challenged traditional gender norms, signaling a growing inclination towards equitable representations. Of notable interest was the acknowledgment of variability in perceptions. Discussions within the workshop revealed personalization of gender bias for participants. Overall, the workshop highlights that gender bias has ill-defined boundaries and a highly subjective interpretation of the concept based on lived experiences, background, education, and other factors. Our workshop not only unveiled the intricacies of biased language and perceptions but also highlighted the need for continued efforts in fostering inclusive dialogue with the community. Our pilot annotation study for annotating gender-biased statements in a comparative annotation setup highlighted challenges with crowdsourcing annotations for this task. One key takeaway was that low-income or rural crowd workers present an additional challenge when dealing with gender bias. While urban audiences can complete such tasks with high accuracy, it is important to think about how we can include minority voices and non-dominant viewpoints for subjective tasks such as gender bias \cite{hada-beyond,stochastic-parrots}. Post-annotation interviews with participants provided valuable context, unveiling a spectrum of challenges encompassing task understanding, virtual training, and language barriers. Women participants, facing connectivity challenges and linguistic complexities, expressed their difficulties with the virtual mode and the formal Hindi used in the sentences. A manual review by native Hindi speakers identified 30\% sentences as misaligned with the cultural context, indicating the necessity for culturally sensitive content. The complexity of translation as also revealed by our experiments, including transliterations and unfamiliar phrases, contributed to misunderstandings among annotators. This annotation study underscores the importance of designing annotation tasks while keeping a variety of audiences in mind. While such tasks have been tested predominantly with the WEIRD or an urban population, not including viewpoints from a rural population can lead to cultural erasure and propagation of hegemonic viewpoints \cite{stochastic-parrots, prabhakaran2022cultural}.  

Overall, our study highlights several challenges faced with mining gender-biased data in the Indian context, subjective understanding of the concept with ill-defined boundaries, and including minority opinions for this task. In the future, it would be interesting to explore a more participatory approach for this task. Works leveraging "games with a purpose" (GWAP) \cite{10.1145/3485447.3512241, 10.1145/1124772.1124784} can be adopted for crowdsourcing generation of gender-biased sentences in target language with cultural nuances, as these methods have shown promise to tap into the tacit knowledge of the crowd. For annotating gender-biased statements a more careful task design is required depending on the target audience, and effort should be made to include minority opinions. It would also be interesting to study the concept of intersectionality introduced by Crenshaw in her foundational work \cite{crenshaw}. She emphasized that social categories like race, gender, and class are interlinked, thereby mutually creating unique dimensions of oppression that are not adequately addressed by frameworks that consider such social categories separately. Drawing on Crenshaw’s work on intersectionality, it is pivotal that the datasets designed to capture gender bias in Indic languages contain sentences that account for bias emerging from these intersecting marginalized identities to authentically capture not just gender bias but its intersection with structural embeddings of caste, religion, and rurality, among others. 

\section{Limitations}
Efforts to mine gender-biased data faced obstacles stemming from the dominance of Anglo-centric content on the internet. Additionally, the growing restrictions on social media further limited our ability to extract diverse and representative datasets. In future, it would be interesting to explore other social media sites like Meta and Twitter if necessary permissions can be obtained. Despite employing various methods, our experiments in mining gender-biased sentences revealed significant difficulties. The low prevalence of biased sentences in the annotated sample suggests challenges in topic selection and the need for domain-specific computational models. The use of translation systems as a bridge between languages revealed substantial limitations. Inability to translate sentences contextually and colloquially hindered our understanding of gender bias in diverse linguistic contexts. The complexities of translation, including transliterations and unfamiliar phrases, contributed to misunderstandings among annotators, underscoring the need for culturally sensitive content. Although the generation of gender bias showed promise, the method exhibited a lack of diversity. In future, it would be interesting to explore GPT generations with more contextual seeds and in-context examples, and changes in the prompt for increasing diversity and including culturally relevant information.
The on-field studies faced challenges in annotating gender-biased statements due to the complex nature of the task. Connectivity issues, linguistic complexities, and cultural misalignments highlighted the difficulties faced by participants. These methodological limitations collectively point to the intricate nature of studying gender bias and emphasize the necessity for continuous refinement of our methods to better capture the complexities inherent in diverse cultural contexts, such as India.
\section{Ethical Considerations}

We use the framework by \citet{bender-friedman-2018-data} to discuss the ethical considerations for our work. 

\begin{itemize}
    \item \textbf{Institutional Review}: All aspects of this research were reviewed and approved by Microsoft Research IRB.
    \item \textbf{Data}: Several publicly available datasets were examined in this study. We also prompt GPT-3.5-Turbo to generate gender biased statements. No personally identifiable information (PII) was collected or distribued in this process. No PII was included in prompt or in-context examples to GPT. 
    \item \textbf{Annotator Demographics}: Annotations for the mining experiments were done by researchers interested in studying gender bias in language technologies. These annotators were from India and had native proficiency in Hindi and English. The annotators had at least an undergraduate degree as their minimum educational qualification. The field studies were conducted with rural, low income women. The demographics are described in section \ref{sec:crowd}. The participants were recruited via a data annotation platform. The pay was adjusted after discussion with the company. The company particularly ensures fair pay (20 times the minimum wage) to low-income communities in India.  
    \item \textbf{Annotation Guidelines}: For the gender bias awareness workshop, as described in section \ref{sec:cs1}, we partner with an organization working in the domain of women's empowerment and gender bias for a long time. All modules were created by this organization. For gender bias annotation workshop we draw from the guidelines described by \citet{hada-etal-2023-fifty}. These guidelines were written in Hindi, and the task was explained verbally during a training session conducted online. Annotators were given detailed instructions, and walk-through of the task, with examples.
    \item \textbf{Impact on Annotators}: For the crowd-sourced annotation study we limited the number of annotations per participant, provided a mix of explicitly, implicitly, and neutrally biased sentences, asked annotators to skip and report instances they were not comfortable annotating, and lastly, encouraged open-dialogue with the workshop facilitator. 
    \item \textbf{Methods}: In this study we explore several methods to mine gender biased data in Hindi. We discuss the challenges and limitations of these methods. We also trained computational models for automatic classification or scoring of gender bias. While these methods can be easily misused, our intent with this study is to highlight the limitations of these methods when used in the context of Global South. 
\end{itemize}
\begin{acks}
This research was supported by Bill and Melinda Gates Foundation (BMGF).
\end{acks}
\bibliographystyle{ACM-Reference-Format}
\bibliography{sample-base}


\begin{thebibliography}{93}


\ifx \showCODEN    \undefined \def \showCODEN     #1{\unskip}     \fi
\ifx \showDOI      \undefined \def \showDOI       #1{#1}\fi
\ifx \showISBNx    \undefined \def \showISBNx     #1{\unskip}     \fi
\ifx \showISBNxiii \undefined \def \showISBNxiii  #1{\unskip}     \fi
\ifx \showISSN     \undefined \def \showISSN      #1{\unskip}     \fi
\ifx \showLCCN     \undefined \def \showLCCN      #1{\unskip}     \fi
\ifx \shownote     \undefined \def \shownote      #1{#1}          \fi
\ifx \showarticletitle \undefined \def \showarticletitle #1{#1}   \fi
\ifx \showURL      \undefined \def \showURL       {\relax}        \fi
\providecommand\bibfield[2]{#2}
\providecommand\bibinfo[2]{#2}
\providecommand\natexlab[1]{#1}
\providecommand\showeprint[2][]{arXiv:#2}

\bibitem[Abraham et~al\mbox{.}(2020)]%
        {abraham-etal-2020-crowdsourcing}
\bibfield{author}{\bibinfo{person}{Basil Abraham}, \bibinfo{person}{Danish Goel}, \bibinfo{person}{Divya Siddarth}, \bibinfo{person}{Kalika Bali}, \bibinfo{person}{Manu Chopra}, \bibinfo{person}{Monojit Choudhury}, \bibinfo{person}{Pratik Joshi}, \bibinfo{person}{Preethi Jyoti}, \bibinfo{person}{Sunayana Sitaram}, {and} \bibinfo{person}{Vivek Seshadri}.} \bibinfo{year}{2020}\natexlab{}.
\newblock \showarticletitle{Crowdsourcing Speech Data for Low-Resource Languages from Low-Income Workers}. In \bibinfo{booktitle}{\emph{Proceedings of the Twelfth Language Resources and Evaluation Conference}}, \bibfield{editor}{\bibinfo{person}{Nicoletta Calzolari}, \bibinfo{person}{Fr{\'e}d{\'e}ric B{\'e}chet}, \bibinfo{person}{Philippe Blache}, \bibinfo{person}{Khalid Choukri}, \bibinfo{person}{Christopher Cieri}, \bibinfo{person}{Thierry Declerck}, \bibinfo{person}{Sara Goggi}, \bibinfo{person}{Hitoshi Isahara}, \bibinfo{person}{Bente Maegaard}, \bibinfo{person}{Joseph Mariani}, \bibinfo{person}{H{\'e}l{\`e}ne Mazo}, \bibinfo{person}{Asuncion Moreno}, \bibinfo{person}{Jan Odijk}, {and} \bibinfo{person}{Stelios Piperidis}} (Eds.). \bibinfo{publisher}{European Language Resources Association}, \bibinfo{address}{Marseille, France}, \bibinfo{pages}{2819--2826}.
\newblock
\showISBNx{979-10-95546-34-4}
\urldef\tempurl%
\url{https://aclanthology.org/2020.lrec-1.343}
\showURL{%
\tempurl}


\bibitem[Ahuja et~al\mbox{.}(2023)]%
        {ahuja-etal-2023-mega}
\bibfield{author}{\bibinfo{person}{Kabir Ahuja}, \bibinfo{person}{Harshita Diddee}, \bibinfo{person}{Rishav Hada}, \bibinfo{person}{Millicent Ochieng}, \bibinfo{person}{Krithika Ramesh}, \bibinfo{person}{Prachi Jain}, \bibinfo{person}{Akshay Nambi}, \bibinfo{person}{Tanuja Ganu}, \bibinfo{person}{Sameer Segal}, \bibinfo{person}{Mohamed Ahmed}, \bibinfo{person}{Kalika Bali}, {and} \bibinfo{person}{Sunayana Sitaram}.} \bibinfo{year}{2023}\natexlab{}.
\newblock \showarticletitle{{MEGA}: Multilingual Evaluation of Generative {AI}}. In \bibinfo{booktitle}{\emph{Proceedings of the 2023 Conference on Empirical Methods in Natural Language Processing}}, \bibfield{editor}{\bibinfo{person}{Houda Bouamor}, \bibinfo{person}{Juan Pino}, {and} \bibinfo{person}{Kalika Bali}} (Eds.). \bibinfo{publisher}{Association for Computational Linguistics}, \bibinfo{address}{Singapore}, \bibinfo{pages}{4232--4267}.
\newblock
\urldef\tempurl%
\url{https://doi.org/10.18653/v1/2023.emnlp-main.258}
\showDOI{\tempurl}


\bibitem[Albalak et~al\mbox{.}(2022)]%
        {albalak2022addressing}
\bibfield{author}{\bibinfo{person}{Alon Albalak}, \bibinfo{person}{Sharon Levy}, {and} \bibinfo{person}{W. Wang}.} \bibinfo{year}{2022}\natexlab{}.
\newblock \showarticletitle{Addressing Issues of Cross-Linguality in Open-Retrieval Question Answering Systems For Emergent Domains}.
\newblock \bibinfo{journal}{\emph{Conference of the European Chapter of the Association for Computational Linguistics}} (\bibinfo{year}{2022}).
\newblock
\urldef\tempurl%
\url{https://doi.org/10.18653/v1/2023.eacl-demo.1}
\showDOI{\tempurl}


\bibitem[Artetxe et~al\mbox{.}(2020)]%
        {artetxe-etal-2020-cross}
\bibfield{author}{\bibinfo{person}{Mikel Artetxe}, \bibinfo{person}{Sebastian Ruder}, {and} \bibinfo{person}{Dani Yogatama}.} \bibinfo{year}{2020}\natexlab{}.
\newblock \showarticletitle{On the Cross-lingual Transferability of Monolingual Representations}. In \bibinfo{booktitle}{\emph{Proceedings of the 58th Annual Meeting of the Association for Computational Linguistics}}, \bibfield{editor}{\bibinfo{person}{Dan Jurafsky}, \bibinfo{person}{Joyce Chai}, \bibinfo{person}{Natalie Schluter}, {and} \bibinfo{person}{Joel Tetreault}} (Eds.). \bibinfo{publisher}{Association for Computational Linguistics}, \bibinfo{address}{Online}, \bibinfo{pages}{4623--4637}.
\newblock
\urldef\tempurl%
\url{https://doi.org/10.18653/v1/2020.acl-main.421}
\showDOI{\tempurl}


\bibitem[Balayn et~al\mbox{.}(2022)]%
        {10.1145/3485447.3512241}
\bibfield{author}{\bibinfo{person}{Agathe Balayn}, \bibinfo{person}{Gaole He}, \bibinfo{person}{Andrea Hu}, \bibinfo{person}{Jie Yang}, {and} \bibinfo{person}{Ujwal Gadiraju}.} \bibinfo{year}{2022}\natexlab{}.
\newblock \showarticletitle{Ready Player One! Eliciting Diverse Knowledge Using A Configurable Game}. In \bibinfo{booktitle}{\emph{Proceedings of the ACM Web Conference 2022}} (Virtual Event, Lyon, France) \emph{(\bibinfo{series}{WWW '22})}. \bibinfo{publisher}{Association for Computing Machinery}, \bibinfo{address}{New York, NY, USA}, \bibinfo{pages}{1709–1719}.
\newblock
\showISBNx{9781450390965}
\urldef\tempurl%
\url{https://doi.org/10.1145/3485447.3512241}
\showDOI{\tempurl}


\bibitem[Barikeri et~al\mbox{.}(2021)]%
        {barikeri-etal-2021-redditbias}
\bibfield{author}{\bibinfo{person}{Soumya Barikeri}, \bibinfo{person}{Anne Lauscher}, \bibinfo{person}{Ivan Vuli{\'c}}, {and} \bibinfo{person}{Goran Glava{\v{s}}}.} \bibinfo{year}{2021}\natexlab{}.
\newblock \showarticletitle{{R}eddit{B}ias: A Real-World Resource for Bias Evaluation and Debiasing of Conversational Language Models}. In \bibinfo{booktitle}{\emph{Proceedings of the 59th Annual Meeting of the Association for Computational Linguistics and the 11th International Joint Conference on Natural Language Processing (Volume 1: Long Papers)}}, \bibfield{editor}{\bibinfo{person}{Chengqing Zong}, \bibinfo{person}{Fei Xia}, \bibinfo{person}{Wenjie Li}, {and} \bibinfo{person}{Roberto Navigli}} (Eds.). \bibinfo{publisher}{Association for Computational Linguistics}, \bibinfo{address}{Online}, \bibinfo{pages}{1941--1955}.
\newblock
\urldef\tempurl%
\url{https://doi.org/10.18653/v1/2021.acl-long.151}
\showDOI{\tempurl}


\bibitem[Bender and Friedman(2018)]%
        {bender-friedman-2018-data}
\bibfield{author}{\bibinfo{person}{Emily~M. Bender} {and} \bibinfo{person}{Batya Friedman}.} \bibinfo{year}{2018}\natexlab{}.
\newblock \showarticletitle{Data Statements for Natural Language Processing: Toward Mitigating System Bias and Enabling Better Science}.
\newblock \bibinfo{journal}{\emph{Transactions of the Association for Computational Linguistics}}  \bibinfo{volume}{6} (\bibinfo{year}{2018}), \bibinfo{pages}{587--604}.
\newblock
\urldef\tempurl%
\url{https://doi.org/10.1162/tacl_a_00041}
\showDOI{\tempurl}


\bibitem[Bender et~al\mbox{.}(2021)]%
        {stochastic-parrots}
\bibfield{author}{\bibinfo{person}{Emily~M. Bender}, \bibinfo{person}{Timnit Gebru}, \bibinfo{person}{Angelina McMillan-Major}, {and} \bibinfo{person}{Shmargaret Shmitchell}.} \bibinfo{year}{2021}\natexlab{}.
\newblock \showarticletitle{On the Dangers of Stochastic Parrots: Can Language Models Be Too Big?}. In \bibinfo{booktitle}{\emph{Proceedings of the 2021 ACM Conference on Fairness, Accountability, and Transparency}} (Virtual Event, Canada) \emph{(\bibinfo{series}{FAccT '21})}. \bibinfo{publisher}{Association for Computing Machinery}, \bibinfo{address}{New York, NY, USA}, \bibinfo{pages}{610–623}.
\newblock
\showISBNx{9781450383097}
\urldef\tempurl%
\url{https://doi.org/10.1145/3442188.3445922}
\showDOI{\tempurl}


\bibitem[Biester et~al\mbox{.}(2022)]%
        {biester-etal-2022-analyzing}
\bibfield{author}{\bibinfo{person}{Laura Biester}, \bibinfo{person}{Vanita Sharma}, \bibinfo{person}{Ashkan Kazemi}, \bibinfo{person}{Naihao Deng}, \bibinfo{person}{Steven Wilson}, {and} \bibinfo{person}{Rada Mihalcea}.} \bibinfo{year}{2022}\natexlab{}.
\newblock \showarticletitle{Analyzing the Effects of Annotator Gender across {NLP} Tasks}. In \bibinfo{booktitle}{\emph{Proceedings of the 1st Workshop on Perspectivist Approaches to NLP @LREC2022}}, \bibfield{editor}{\bibinfo{person}{Gavin Abercrombie}, \bibinfo{person}{Valerio Basile}, \bibinfo{person}{Sara Tonelli}, \bibinfo{person}{Verena Rieser}, {and} \bibinfo{person}{Alexandra Uma}} (Eds.). \bibinfo{publisher}{European Language Resources Association}, \bibinfo{address}{Marseille, France}, \bibinfo{pages}{10--19}.
\newblock
\urldef\tempurl%
\url{https://aclanthology.org/2022.nlperspectives-1.2}
\showURL{%
\tempurl}


\bibitem[Birhane et~al\mbox{.}(2022)]%
        {10.1145/3551624.3555290}
\bibfield{author}{\bibinfo{person}{Abeba Birhane}, \bibinfo{person}{William Isaac}, \bibinfo{person}{Vinodkumar Prabhakaran}, \bibinfo{person}{Mark Diaz}, \bibinfo{person}{Madeleine~Clare Elish}, \bibinfo{person}{Iason Gabriel}, {and} \bibinfo{person}{Shakir Mohamed}.} \bibinfo{year}{2022}\natexlab{}.
\newblock \showarticletitle{Power to the People? Opportunities and Challenges for Participatory AI}. In \bibinfo{booktitle}{\emph{Proceedings of the 2nd ACM Conference on Equity and Access in Algorithms, Mechanisms, and Optimization}} (<conf-loc>, <city>Arlington</city>, <state>VA</state>, <country>USA</country>, </conf-loc>) \emph{(\bibinfo{series}{EAAMO '22})}. \bibinfo{publisher}{Association for Computing Machinery}, \bibinfo{address}{New York, NY, USA}, Article \bibinfo{articleno}{6}, \bibinfo{numpages}{8}~pages.
\newblock
\showISBNx{9781450394772}
\urldef\tempurl%
\url{https://doi.org/10.1145/3551624.3555290}
\showDOI{\tempurl}


\bibitem[Blodgett et~al\mbox{.}(2020)]%
        {blodgett-etal-2020-language}
\bibfield{author}{\bibinfo{person}{Su~Lin Blodgett}, \bibinfo{person}{Solon Barocas}, \bibinfo{person}{Hal Daum{\'e}~III}, {and} \bibinfo{person}{Hanna Wallach}.} \bibinfo{year}{2020}\natexlab{}.
\newblock \showarticletitle{Language (Technology) is Power: A Critical Survey of {``}Bias{''} in {NLP}}. In \bibinfo{booktitle}{\emph{Proceedings of the 58th Annual Meeting of the Association for Computational Linguistics}}, \bibfield{editor}{\bibinfo{person}{Dan Jurafsky}, \bibinfo{person}{Joyce Chai}, \bibinfo{person}{Natalie Schluter}, {and} \bibinfo{person}{Joel Tetreault}} (Eds.). \bibinfo{publisher}{Association for Computational Linguistics}, \bibinfo{address}{Online}, \bibinfo{pages}{5454--5476}.
\newblock
\urldef\tempurl%
\url{https://doi.org/10.18653/v1/2020.acl-main.485}
\showDOI{\tempurl}


\bibitem[Bolukbasi et~al\mbox{.}(2016)]%
        {DBLP:conf/nips/BolukbasiCZSK16}
\bibfield{author}{\bibinfo{person}{Tolga Bolukbasi}, \bibinfo{person}{Kai{-}Wei Chang}, \bibinfo{person}{James~Y. Zou}, \bibinfo{person}{Venkatesh Saligrama}, {and} \bibinfo{person}{Adam~Tauman Kalai}.} \bibinfo{year}{2016}\natexlab{}.
\newblock \showarticletitle{Man is to Computer Programmer as Woman is to Homemaker? Debiasing Word Embeddings}. In \bibinfo{booktitle}{\emph{Advances in Neural Information Processing Systems 29: Annual Conference on Neural Information Processing Systems 2016, December 5-10, 2016, Barcelona, Spain}}, \bibfield{editor}{\bibinfo{person}{Daniel~D. Lee}, \bibinfo{person}{Masashi Sugiyama}, \bibinfo{person}{Ulrike von Luxburg}, \bibinfo{person}{Isabelle Guyon}, {and} \bibinfo{person}{Roman Garnett}} (Eds.). \bibinfo{pages}{4349--4357}.
\newblock
\urldef\tempurl%
\url{https://proceedings.neurips.cc/paper/2016/hash/a486cd07e4ac3d270571622f4f316ec5-Abstract.html}
\showURL{%
\tempurl}


\bibitem[Braff and Nelson(2022)]%
        {braff2022chapter}
\bibfield{author}{\bibinfo{person}{Lara Braff} {and} \bibinfo{person}{Katie Nelson}.} \bibinfo{year}{2022}\natexlab{}.
\newblock \showarticletitle{Chapter 15: The Global North: Introducing the Region}.
\newblock \bibinfo{journal}{\emph{Gendered Lives}} (\bibinfo{year}{2022}).
\newblock


\bibitem[Buschek et~al\mbox{.}(2021)]%
        {10.1145/3411764.3445372}
\bibfield{author}{\bibinfo{person}{Daniel Buschek}, \bibinfo{person}{Martin Z\"{u}rn}, {and} \bibinfo{person}{Malin Eiband}.} \bibinfo{year}{2021}\natexlab{}.
\newblock \showarticletitle{The Impact of Multiple Parallel Phrase Suggestions on Email Input and Composition Behaviour of Native and Non-Native English Writers}. In \bibinfo{booktitle}{\emph{Proceedings of the 2021 CHI Conference on Human Factors in Computing Systems}} (<conf-loc>, <city>Yokohama</city>, <country>Japan</country>, </conf-loc>) \emph{(\bibinfo{series}{CHI '21})}. \bibinfo{publisher}{Association for Computing Machinery}, \bibinfo{address}{New York, NY, USA}, Article \bibinfo{articleno}{732}, \bibinfo{numpages}{13}~pages.
\newblock
\showISBNx{9781450380966}
\urldef\tempurl%
\url{https://doi.org/10.1145/3411764.3445372}
\showDOI{\tempurl}


\bibitem[Caliskan et~al\mbox{.}(2017)]%
        {doi:10.1126/science.aal4230}
\bibfield{author}{\bibinfo{person}{Aylin Caliskan}, \bibinfo{person}{Joanna~J. Bryson}, {and} \bibinfo{person}{Arvind Narayanan}.} \bibinfo{year}{2017}\natexlab{}.
\newblock \showarticletitle{Semantics derived automatically from language corpora contain human-like biases}.
\newblock \bibinfo{journal}{\emph{Science}} \bibinfo{volume}{356}, \bibinfo{number}{6334} (\bibinfo{year}{2017}), \bibinfo{pages}{183--186}.
\newblock
\urldef\tempurl%
\url{https://doi.org/10.1126/science.aal4230}
\showDOI{\tempurl}
\showeprint{https://www.science.org/doi/pdf/10.1126/science.aal4230}


\bibitem[Choi et~al\mbox{.}(2023)]%
        {choi-etal-2023-llms}
\bibfield{author}{\bibinfo{person}{Minje Choi}, \bibinfo{person}{Jiaxin Pei}, \bibinfo{person}{Sagar Kumar}, \bibinfo{person}{Chang Shu}, {and} \bibinfo{person}{David Jurgens}.} \bibinfo{year}{2023}\natexlab{}.
\newblock \showarticletitle{Do {LLM}s Understand Social Knowledge? Evaluating the Sociability of Large Language Models with {S}oc{KET} Benchmark}. In \bibinfo{booktitle}{\emph{Proceedings of the 2023 Conference on Empirical Methods in Natural Language Processing}}, \bibfield{editor}{\bibinfo{person}{Houda Bouamor}, \bibinfo{person}{Juan Pino}, {and} \bibinfo{person}{Kalika Bali}} (Eds.). \bibinfo{publisher}{Association for Computational Linguistics}, \bibinfo{address}{Singapore}, \bibinfo{pages}{11370--11403}.
\newblock
\urldef\tempurl%
\url{https://doi.org/10.18653/v1/2023.emnlp-main.699}
\showDOI{\tempurl}


\bibitem[Chopra et~al\mbox{.}(2019)]%
        {chopra_karya}
\bibfield{author}{\bibinfo{person}{Manu Chopra}, \bibinfo{person}{Indrani Medhi~Thies}, \bibinfo{person}{Joyojeet Pal}, \bibinfo{person}{Colin Scott}, \bibinfo{person}{William Thies}, {and} \bibinfo{person}{Vivek Seshadri}.} \bibinfo{year}{2019}\natexlab{}.
\newblock \showarticletitle{Exploring Crowdsourced Work in Low-Resource Settings}. In \bibinfo{booktitle}{\emph{Proceedings of the 2019 CHI Conference on Human Factors in Computing Systems}} (Glasgow, Scotland Uk) \emph{(\bibinfo{series}{CHI '19})}. \bibinfo{publisher}{Association for Computing Machinery}, \bibinfo{address}{New York, NY, USA}, \bibinfo{pages}{1–13}.
\newblock
\showISBNx{9781450359702}
\urldef\tempurl%
\url{https://doi.org/10.1145/3290605.3300611}
\showDOI{\tempurl}


\bibitem[Crenshaw(1991)]%
        {crenshaw}
\bibfield{author}{\bibinfo{person}{Kimberle Crenshaw}.} \bibinfo{year}{1991}\natexlab{}.
\newblock \showarticletitle{Mapping the Margins: Intersectionality, Identity Politics, and Violence against Women of Color}.
\newblock \bibinfo{journal}{\emph{Stanford Law Review}} \bibinfo{volume}{43}, \bibinfo{number}{6} (\bibinfo{year}{1991}), \bibinfo{pages}{1241--1299}.
\newblock
\showISSN{00389765}
\urldef\tempurl%
\url{http://www.jstor.org/stable/1229039}
\showURL{%
\tempurl}


\bibitem[Davidson et~al\mbox{.}(2023)]%
        {davidson2023platform}
\bibfield{author}{\bibinfo{person}{Brittany~I Davidson}, \bibinfo{person}{Darja Wischerath}, \bibinfo{person}{Daniel Racek}, \bibinfo{person}{Douglas~A Parry}, \bibinfo{person}{Emily Godwin}, \bibinfo{person}{Joanne Hinds}, \bibinfo{person}{Dirk van~der Linden}, \bibinfo{person}{Jonathan~F Roscoe}, \bibinfo{person}{Laura Ayravainen}, {and} \bibinfo{person}{Alicia~G Cork}.} \bibinfo{year}{2023}\natexlab{}.
\newblock \showarticletitle{Platform-controlled social media APIs threaten Open Science}.
\newblock \bibinfo{journal}{\emph{Nature Human Behaviour}} (\bibinfo{year}{2023}), \bibinfo{pages}{1--4}.
\newblock


\bibitem[Dev et~al\mbox{.}(2022)]%
        {dev-etal-2022-measures}
\bibfield{author}{\bibinfo{person}{Sunipa Dev}, \bibinfo{person}{Emily Sheng}, \bibinfo{person}{Jieyu Zhao}, \bibinfo{person}{Aubrie Amstutz}, \bibinfo{person}{Jiao Sun}, \bibinfo{person}{Yu Hou}, \bibinfo{person}{Mattie Sanseverino}, \bibinfo{person}{Jiin Kim}, \bibinfo{person}{Akihiro Nishi}, \bibinfo{person}{Nanyun Peng}, {and} \bibinfo{person}{Kai-Wei Chang}.} \bibinfo{year}{2022}\natexlab{}.
\newblock \showarticletitle{On Measures of Biases and Harms in {NLP}}. In \bibinfo{booktitle}{\emph{Findings of the Association for Computational Linguistics: AACL-IJCNLP 2022}}, \bibfield{editor}{\bibinfo{person}{Yulan He}, \bibinfo{person}{Heng Ji}, \bibinfo{person}{Sujian Li}, \bibinfo{person}{Yang Liu}, {and} \bibinfo{person}{Chua-Hui Chang}} (Eds.). \bibinfo{publisher}{Association for Computational Linguistics}, \bibinfo{address}{Online only}, \bibinfo{pages}{246--267}.
\newblock
\urldef\tempurl%
\url{https://aclanthology.org/2022.findings-aacl.24}
\showURL{%
\tempurl}


\bibitem[Devinney et~al\mbox{.}(2022)]%
        {10.1145/3531146.3534627}
\bibfield{author}{\bibinfo{person}{Hannah Devinney}, \bibinfo{person}{Jenny Bj\"{o}rklund}, {and} \bibinfo{person}{Henrik Bj\"{o}rklund}.} \bibinfo{year}{2022}\natexlab{}.
\newblock \showarticletitle{Theories of “Gender” in NLP Bias Research}. In \bibinfo{booktitle}{\emph{Proceedings of the 2022 ACM Conference on Fairness, Accountability, and Transparency}} (Seoul, Republic of Korea) \emph{(\bibinfo{series}{FAccT '22})}. \bibinfo{publisher}{Association for Computing Machinery}, \bibinfo{address}{New York, NY, USA}, \bibinfo{pages}{2083–2102}.
\newblock
\showISBNx{9781450393522}
\urldef\tempurl%
\url{https://doi.org/10.1145/3531146.3534627}
\showDOI{\tempurl}


\bibitem[Devlin et~al\mbox{.}(2019a)]%
        {devlin-etal-2019-bert}
\bibfield{author}{\bibinfo{person}{Jacob Devlin}, \bibinfo{person}{Ming-Wei Chang}, \bibinfo{person}{Kenton Lee}, {and} \bibinfo{person}{Kristina Toutanova}.} \bibinfo{year}{2019}\natexlab{a}.
\newblock \showarticletitle{{BERT}: Pre-training of Deep Bidirectional Transformers for Language Understanding}. In \bibinfo{booktitle}{\emph{Proceedings of the 2019 Conference of the North {A}merican Chapter of the Association for Computational Linguistics: Human Language Technologies, Volume 1 (Long and Short Papers)}}, \bibfield{editor}{\bibinfo{person}{Jill Burstein}, \bibinfo{person}{Christy Doran}, {and} \bibinfo{person}{Thamar Solorio}} (Eds.). \bibinfo{publisher}{Association for Computational Linguistics}, \bibinfo{address}{Minneapolis, Minnesota}, \bibinfo{pages}{4171--4186}.
\newblock
\urldef\tempurl%
\url{https://doi.org/10.18653/v1/N19-1423}
\showDOI{\tempurl}


\bibitem[Devlin et~al\mbox{.}(2019b)]%
        {Devlin2019BERTPO}
\bibfield{author}{\bibinfo{person}{Jacob Devlin}, \bibinfo{person}{Ming-Wei Chang}, \bibinfo{person}{Kenton Lee}, {and} \bibinfo{person}{Kristina Toutanova}.} \bibinfo{year}{2019}\natexlab{b}.
\newblock \showarticletitle{BERT: Pre-training of Deep Bidirectional Transformers for Language Understanding}. In \bibinfo{booktitle}{\emph{North American Chapter of the Association for Computational Linguistics}}.
\newblock
\urldef\tempurl%
\url{https://api.semanticscholar.org/CorpusID:52967399}
\showURL{%
\tempurl}


\bibitem[Difallah et~al\mbox{.}({[n.\,d.]})]%
        {Mechanic29:online}
\bibfield{author}{\bibinfo{person}{Djellel Difallah}, \bibinfo{person}{Elena Filatova}, {and} \bibinfo{person}{Panos Ipeirotis}.} \bibinfo{year}{[n.\,d.]}\natexlab{}.
\newblock \bibinfo{title}{Mechanical Turk Surveys}.
\newblock \bibinfo{howpublished}{\url{https://demographics.mturk-tracker.com/}}.
\newblock
\newblock
\shownote{(Accessed on 09/11/2023)}.


\bibitem[Difallah et~al\mbox{.}(2018)]%
        {demo_dyn_mtutk}
\bibfield{author}{\bibinfo{person}{Djellel Difallah}, \bibinfo{person}{Elena Filatova}, {and} \bibinfo{person}{Panos Ipeirotis}.} \bibinfo{year}{2018}\natexlab{}.
\newblock \showarticletitle{Demographics and Dynamics of Mechanical Turk Workers} \emph{(\bibinfo{series}{WSDM '18})}. \bibinfo{publisher}{Association for Computing Machinery}, \bibinfo{address}{New York, NY, USA}, \bibinfo{pages}{135–143}.
\newblock
\showISBNx{9781450355810}
\urldef\tempurl%
\url{https://doi.org/10.1145/3159652.3159661}
\showDOI{\tempurl}


\bibitem[Doddapaneni et~al\mbox{.}(2023)]%
        {doddapaneni-etal-2023-towards}
\bibfield{author}{\bibinfo{person}{Sumanth Doddapaneni}, \bibinfo{person}{Rahul Aralikatte}, \bibinfo{person}{Gowtham Ramesh}, \bibinfo{person}{Shreya Goyal}, \bibinfo{person}{Mitesh~M. Khapra}, \bibinfo{person}{Anoop Kunchukuttan}, {and} \bibinfo{person}{Pratyush Kumar}.} \bibinfo{year}{2023}\natexlab{}.
\newblock \showarticletitle{Towards Leaving No {I}ndic Language Behind: Building Monolingual Corpora, Benchmark and Models for {I}ndic Languages}. In \bibinfo{booktitle}{\emph{Proceedings of the 61st Annual Meeting of the Association for Computational Linguistics (Volume 1: Long Papers)}}, \bibfield{editor}{\bibinfo{person}{Anna Rogers}, \bibinfo{person}{Jordan Boyd-Graber}, {and} \bibinfo{person}{Naoaki Okazaki}} (Eds.). \bibinfo{publisher}{Association for Computational Linguistics}, \bibinfo{address}{Toronto, Canada}, \bibinfo{pages}{12402--12426}.
\newblock
\urldef\tempurl%
\url{https://doi.org/10.18653/v1/2023.acl-long.693}
\showDOI{\tempurl}


\bibitem[Draws et~al\mbox{.}(2021)]%
        {draws2021checklist}
\bibfield{author}{\bibinfo{person}{Tim Draws}, \bibinfo{person}{Alisa Rieger}, \bibinfo{person}{Oana Inel}, \bibinfo{person}{Ujwal Gadiraju}, {and} \bibinfo{person}{Nava Tintarev}.} \bibinfo{year}{2021}\natexlab{}.
\newblock \showarticletitle{A checklist to combat cognitive biases in crowdsourcing}. In \bibinfo{booktitle}{\emph{Proceedings of the AAAI conference on human computation and crowdsourcing}}, Vol.~\bibinfo{volume}{9}. \bibinfo{pages}{48--59}.
\newblock


\bibitem[F\"{a}rber et~al\mbox{.}(2020)]%
        {10.1145/3340531.3412876}
\bibfield{author}{\bibinfo{person}{Michael F\"{a}rber}, \bibinfo{person}{Victoria Burkard}, \bibinfo{person}{Adam Jatowt}, {and} \bibinfo{person}{Sora Lim}.} \bibinfo{year}{2020}\natexlab{}.
\newblock \showarticletitle{A Multidimensional Dataset Based on Crowdsourcing for Analyzing and Detecting News Bias}. In \bibinfo{booktitle}{\emph{Proceedings of the 29th ACM International Conference on Information \& Knowledge Management}} (Virtual Event, Ireland) \emph{(\bibinfo{series}{CIKM '20})}. \bibinfo{publisher}{Association for Computing Machinery}, \bibinfo{address}{New York, NY, USA}, \bibinfo{pages}{3007–3014}.
\newblock
\showISBNx{9781450368599}
\urldef\tempurl%
\url{https://doi.org/10.1145/3340531.3412876}
\showDOI{\tempurl}


\bibitem[Feng et~al\mbox{.}(2022)]%
        {feng-etal-2022-language}
\bibfield{author}{\bibinfo{person}{Fangxiaoyu Feng}, \bibinfo{person}{Yinfei Yang}, \bibinfo{person}{Daniel Cer}, \bibinfo{person}{Naveen Arivazhagan}, {and} \bibinfo{person}{Wei Wang}.} \bibinfo{year}{2022}\natexlab{}.
\newblock \showarticletitle{Language-agnostic {BERT} Sentence Embedding}. In \bibinfo{booktitle}{\emph{Proceedings of the 60th Annual Meeting of the Association for Computational Linguistics (Volume 1: Long Papers)}}, \bibfield{editor}{\bibinfo{person}{Smaranda Muresan}, \bibinfo{person}{Preslav Nakov}, {and} \bibinfo{person}{Aline Villavicencio}} (Eds.). \bibinfo{publisher}{Association for Computational Linguistics}, \bibinfo{address}{Dublin, Ireland}, \bibinfo{pages}{878--891}.
\newblock
\urldef\tempurl%
\url{https://doi.org/10.18653/v1/2022.acl-long.62}
\showDOI{\tempurl}


\bibitem[Flynn and Marley(2014)]%
        {RePEc:elg:eechap:14820_8}
\bibfield{author}{\bibinfo{person}{T.N. Flynn} {and} \bibinfo{person}{A.A.J. Marley}.} \bibinfo{year}{2014}\natexlab{}.
\newblock \showarticletitle{{Best-worst scaling: theory and methods}}.
\newblock In \bibinfo{booktitle}{\emph{{Handbook of Choice Modelling}}}, \bibfield{editor}{\bibinfo{person}{Stephane Hess} {and} \bibinfo{person}{Andrew Daly}} (Eds.). \bibinfo{publisher}{Edward Elgar Publishing}, Chapter~8, \bibinfo{pages}{178--201}.
\newblock
\urldef\tempurl%
\url{https://ideas.repec.org/h/elg/eechap/14820_8.html}
\showURL{%
\tempurl}


\bibitem[for Economic Co-operation and (OECD)(2018)]%
        {organisation2018bridging}
\bibfield{author}{\bibinfo{person}{Organisation for Economic Co-operation} {and} \bibinfo{person}{Development (OECD)}.} \bibinfo{year}{2018}\natexlab{}.
\newblock \showarticletitle{Bridging the digital gender divide: Include, upskill, innovate}.
\newblock \bibinfo{journal}{\emph{OECD}} (\bibinfo{year}{2018}).
\newblock


\bibitem[Founta et~al\mbox{.}(2018)]%
        {Founta}
\bibfield{author}{\bibinfo{person}{Antigoni Founta}, \bibinfo{person}{Constantinos Djouvas}, \bibinfo{person}{Despoina Chatzakou}, \bibinfo{person}{Ilias Leontiadis}, \bibinfo{person}{Jeremy Blackburn}, \bibinfo{person}{Gianluca Stringhini}, \bibinfo{person}{Athena Vakali}, \bibinfo{person}{Michael Sirivianos}, {and} \bibinfo{person}{Nicolas Kourtellis}.} \bibinfo{year}{2018}\natexlab{}.
\newblock \showarticletitle{Large Scale Crowdsourcing and Characterization of Twitter Abusive Behavior}.
\newblock \bibinfo{journal}{\emph{Proceedings of the International AAAI Conference on Web and Social Media}} \bibinfo{volume}{12}, \bibinfo{number}{1} (\bibinfo{date}{Jun.} \bibinfo{year}{2018}).
\newblock
\urldef\tempurl%
\url{https://doi.org/10.1609/icwsm.v12i1.14991}
\showDOI{\tempurl}


\bibitem[Gala et~al\mbox{.}(2023)]%
        {gala2023indictrans}
\bibfield{author}{\bibinfo{person}{Jay Gala}, \bibinfo{person}{Pranjal~A Chitale}, \bibinfo{person}{A~K Raghavan}, \bibinfo{person}{Varun Gumma}, \bibinfo{person}{Sumanth Doddapaneni}, \bibinfo{person}{Aswanth~Kumar M}, \bibinfo{person}{Janki~Atul Nawale}, \bibinfo{person}{Anupama Sujatha}, \bibinfo{person}{Ratish Puduppully}, \bibinfo{person}{Vivek Raghavan}, \bibinfo{person}{Pratyush Kumar}, \bibinfo{person}{Mitesh~M Khapra}, \bibinfo{person}{Raj Dabre}, {and} \bibinfo{person}{Anoop Kunchukuttan}.} \bibinfo{year}{2023}\natexlab{}.
\newblock \showarticletitle{IndicTrans2: Towards High-Quality and Accessible Machine Translation Models for all 22 Scheduled Indian Languages}.
\newblock \bibinfo{journal}{\emph{Transactions on Machine Learning Research}} (\bibinfo{year}{2023}).
\newblock
\showISSN{2835-8856}
\urldef\tempurl%
\url{https://openreview.net/forum?id=vfT4YuzAYA}
\showURL{%
\tempurl}


\bibitem[Hada et~al\mbox{.}(2023a)]%
        {hada-beyond}
\bibfield{author}{\bibinfo{person}{Rishav Hada}, \bibinfo{person}{Amir Ebrahimi~Fard}, \bibinfo{person}{Sarah Shugars}, \bibinfo{person}{Federico Bianchi}, \bibinfo{person}{Patricia Rossini}, \bibinfo{person}{Dirk Hovy}, \bibinfo{person}{Rebekah Tromble}, {and} \bibinfo{person}{Nava Tintarev}.} \bibinfo{year}{2023}\natexlab{a}.
\newblock \showarticletitle{Beyond Digital "Echo Chambers": The Role of Viewpoint Diversity in Political Discussion}. In \bibinfo{booktitle}{\emph{Proceedings of the Sixteenth ACM International Conference on Web Search and Data Mining}} (<conf-loc>, <city>Singapore</city>, <country>Singapore</country>, </conf-loc>) \emph{(\bibinfo{series}{WSDM '23})}. \bibinfo{publisher}{Association for Computing Machinery}, \bibinfo{address}{New York, NY, USA}, \bibinfo{pages}{33–41}.
\newblock
\showISBNx{9781450394079}
\urldef\tempurl%
\url{https://doi.org/10.1145/3539597.3570487}
\showDOI{\tempurl}


\bibitem[Hada et~al\mbox{.}(2023b)]%
        {hada-etal-2023-fifty}
\bibfield{author}{\bibinfo{person}{Rishav Hada}, \bibinfo{person}{Agrima Seth}, \bibinfo{person}{Harshita Diddee}, {and} \bibinfo{person}{Kalika Bali}.} \bibinfo{year}{2023}\natexlab{b}.
\newblock \showarticletitle{{``}Fifty Shades of Bias{''}: Normative Ratings of Gender Bias in {GPT} Generated {E}nglish Text}. In \bibinfo{booktitle}{\emph{Proceedings of the 2023 Conference on Empirical Methods in Natural Language Processing}}, \bibfield{editor}{\bibinfo{person}{Houda Bouamor}, \bibinfo{person}{Juan Pino}, {and} \bibinfo{person}{Kalika Bali}} (Eds.). \bibinfo{publisher}{Association for Computational Linguistics}, \bibinfo{address}{Singapore}, \bibinfo{pages}{1862--1876}.
\newblock
\urldef\tempurl%
\url{https://doi.org/10.18653/v1/2023.emnlp-main.115}
\showDOI{\tempurl}


\bibitem[Hada et~al\mbox{.}(2021)]%
        {hada-etal-2021-ruddit}
\bibfield{author}{\bibinfo{person}{Rishav Hada}, \bibinfo{person}{Sohi Sudhir}, \bibinfo{person}{Pushkar Mishra}, \bibinfo{person}{Helen Yannakoudakis}, \bibinfo{person}{Saif~M. Mohammad}, {and} \bibinfo{person}{Ekaterina Shutova}.} \bibinfo{year}{2021}\natexlab{}.
\newblock \showarticletitle{Ruddit: {N}orms of Offensiveness for {E}nglish {R}eddit Comments}. In \bibinfo{booktitle}{\emph{Proceedings of the 59th Annual Meeting of the Association for Computational Linguistics and the 11th International Joint Conference on Natural Language Processing (Volume 1: Long Papers)}}, \bibfield{editor}{\bibinfo{person}{Chengqing Zong}, \bibinfo{person}{Fei Xia}, \bibinfo{person}{Wenjie Li}, {and} \bibinfo{person}{Roberto Navigli}} (Eds.). \bibinfo{publisher}{Association for Computational Linguistics}, \bibinfo{address}{Online}, \bibinfo{pages}{2700--2717}.
\newblock
\urldef\tempurl%
\url{https://doi.org/10.18653/v1/2021.acl-long.210}
\showDOI{\tempurl}


\bibitem[Hu et~al\mbox{.}(2022)]%
        {hu2022lora}
\bibfield{author}{\bibinfo{person}{Edward~J Hu}, \bibinfo{person}{yelong shen}, \bibinfo{person}{Phillip Wallis}, \bibinfo{person}{Zeyuan Allen-Zhu}, \bibinfo{person}{Yuanzhi Li}, \bibinfo{person}{Shean Wang}, \bibinfo{person}{Lu Wang}, {and} \bibinfo{person}{Weizhu Chen}.} \bibinfo{year}{2022}\natexlab{}.
\newblock \showarticletitle{Lo{RA}: Low-Rank Adaptation of Large Language Models}. In \bibinfo{booktitle}{\emph{International Conference on Learning Representations}}.
\newblock
\urldef\tempurl%
\url{https://openreview.net/forum?id=nZeVKeeFYf9}
\showURL{%
\tempurl}


\bibitem[Javed et~al\mbox{.}(2023)]%
        {10.1609/aaai.v37i11.26521}
\bibfield{author}{\bibinfo{person}{Tahir Javed}, \bibinfo{person}{Kaushal Bhogale}, \bibinfo{person}{Abhigyan Raman}, \bibinfo{person}{Pratyush Kumar}, \bibinfo{person}{Anoop Kunchukuttan}, {and} \bibinfo{person}{Mitesh~M. Khapra}.} \bibinfo{year}{2023}\natexlab{}.
\newblock \showarticletitle{IndicSUPERB: A Speech Processing Universal Performance Benchmark for Indian Languages}. In \bibinfo{booktitle}{\emph{Proceedings of the Thirty-Seventh AAAI Conference on Artificial Intelligence and Thirty-Fifth Conference on Innovative Applications of Artificial Intelligence and Thirteenth Symposium on Educational Advances in Artificial Intelligence}} \emph{(\bibinfo{series}{AAAI'23/IAAI'23/EAAI'23})}. \bibinfo{publisher}{AAAI Press}, Article \bibinfo{articleno}{1452}, \bibinfo{numpages}{9}~pages.
\newblock
\showISBNx{978-1-57735-880-0}
\urldef\tempurl%
\url{https://doi.org/10.1609/aaai.v37i11.26521}
\showDOI{\tempurl}


\bibitem[Joshi et~al\mbox{.}(2019)]%
        {joshi-etal-2019-unsung}
\bibfield{author}{\bibinfo{person}{Pratik Joshi}, \bibinfo{person}{Christain Barnes}, \bibinfo{person}{Sebastin Santy}, \bibinfo{person}{Simran Khanuja}, \bibinfo{person}{Sanket Shah}, \bibinfo{person}{Anirudh Srinivasan}, \bibinfo{person}{Satwik Bhattamishra}, \bibinfo{person}{Sunayana Sitaram}, \bibinfo{person}{Monojit Choudhury}, {and} \bibinfo{person}{Kalika Bali}.} \bibinfo{year}{2019}\natexlab{}.
\newblock \showarticletitle{Unsung Challenges of Building and Deploying Language Technologies for Low Resource Language Communities}. In \bibinfo{booktitle}{\emph{Proceedings of the 16th International Conference on Natural Language Processing}}. \bibinfo{publisher}{NLP Association of India}, \bibinfo{address}{International Institute of Information Technology, Hyderabad, India}, \bibinfo{pages}{211--219}.
\newblock
\urldef\tempurl%
\url{https://aclanthology.org/2019.icon-1.25}
\showURL{%
\tempurl}


\bibitem[Joshi et~al\mbox{.}(2020)]%
        {joshi-etal-2020-state}
\bibfield{author}{\bibinfo{person}{Pratik Joshi}, \bibinfo{person}{Sebastin Santy}, \bibinfo{person}{Amar Budhiraja}, \bibinfo{person}{Kalika Bali}, {and} \bibinfo{person}{Monojit Choudhury}.} \bibinfo{year}{2020}\natexlab{}.
\newblock \showarticletitle{The State and Fate of Linguistic Diversity and Inclusion in the {NLP} World}. In \bibinfo{booktitle}{\emph{Proceedings of the 58th Annual Meeting of the Association for Computational Linguistics}}, \bibfield{editor}{\bibinfo{person}{Dan Jurafsky}, \bibinfo{person}{Joyce Chai}, \bibinfo{person}{Natalie Schluter}, {and} \bibinfo{person}{Joel Tetreault}} (Eds.). \bibinfo{publisher}{Association for Computational Linguistics}, \bibinfo{address}{Online}, \bibinfo{pages}{6282--6293}.
\newblock
\urldef\tempurl%
\url{https://doi.org/10.18653/v1/2020.acl-main.560}
\showDOI{\tempurl}


\bibitem[Kakwani et~al\mbox{.}(2020)]%
        {kakwani-etal-2020-indicnlpsuite}
\bibfield{author}{\bibinfo{person}{Divyanshu Kakwani}, \bibinfo{person}{Anoop Kunchukuttan}, \bibinfo{person}{Satish Golla}, \bibinfo{person}{Gokul N.C.}, \bibinfo{person}{Avik Bhattacharyya}, \bibinfo{person}{Mitesh~M. Khapra}, {and} \bibinfo{person}{Pratyush Kumar}.} \bibinfo{year}{2020}\natexlab{}.
\newblock \showarticletitle{{I}ndic{NLPS}uite: Monolingual Corpora, Evaluation Benchmarks and Pre-trained Multilingual Language Models for {I}ndian Languages}. In \bibinfo{booktitle}{\emph{Findings of the Association for Computational Linguistics: EMNLP 2020}}, \bibfield{editor}{\bibinfo{person}{Trevor Cohn}, \bibinfo{person}{Yulan He}, {and} \bibinfo{person}{Yang Liu}} (Eds.). \bibinfo{publisher}{Association for Computational Linguistics}, \bibinfo{address}{Online}, \bibinfo{pages}{4948--4961}.
\newblock
\urldef\tempurl%
\url{https://doi.org/10.18653/v1/2020.findings-emnlp.445}
\showDOI{\tempurl}


\bibitem[Kaneko et~al\mbox{.}(2022a)]%
        {kaneko-etal-2022-debiasing}
\bibfield{author}{\bibinfo{person}{Masahiro Kaneko}, \bibinfo{person}{Danushka Bollegala}, {and} \bibinfo{person}{Naoaki Okazaki}.} \bibinfo{year}{2022}\natexlab{a}.
\newblock \showarticletitle{Debiasing Isn{'}t Enough! {--} on the Effectiveness of Debiasing {MLM}s and Their Social Biases in Downstream Tasks}. In \bibinfo{booktitle}{\emph{Proceedings of the 29th International Conference on Computational Linguistics}}, \bibfield{editor}{\bibinfo{person}{Nicoletta Calzolari}, \bibinfo{person}{Chu-Ren Huang}, \bibinfo{person}{Hansaem Kim}, \bibinfo{person}{James Pustejovsky}, \bibinfo{person}{Leo Wanner}, \bibinfo{person}{Key-Sun Choi}, \bibinfo{person}{Pum-Mo Ryu}, \bibinfo{person}{Hsin-Hsi Chen}, \bibinfo{person}{Lucia Donatelli}, \bibinfo{person}{Heng Ji}, \bibinfo{person}{Sadao Kurohashi}, \bibinfo{person}{Patrizia Paggio}, \bibinfo{person}{Nianwen Xue}, \bibinfo{person}{Seokhwan Kim}, \bibinfo{person}{Younggyun Hahm}, \bibinfo{person}{Zhong He}, \bibinfo{person}{Tony~Kyungil Lee}, \bibinfo{person}{Enrico Santus}, \bibinfo{person}{Francis Bond}, {and} \bibinfo{person}{Seung-Hoon Na}} (Eds.). \bibinfo{publisher}{International Committee on Computational Linguistics},
  \bibinfo{address}{Gyeongju, Republic of Korea}, \bibinfo{pages}{1299--1310}.
\newblock
\urldef\tempurl%
\url{https://aclanthology.org/2022.coling-1.111}
\showURL{%
\tempurl}


\bibitem[Kaneko et~al\mbox{.}(2022b)]%
        {kaneko-etal-2022-gender}
\bibfield{author}{\bibinfo{person}{Masahiro Kaneko}, \bibinfo{person}{Aizhan Imankulova}, \bibinfo{person}{Danushka Bollegala}, {and} \bibinfo{person}{Naoaki Okazaki}.} \bibinfo{year}{2022}\natexlab{b}.
\newblock \showarticletitle{Gender Bias in Masked Language Models for Multiple Languages}. In \bibinfo{booktitle}{\emph{Proceedings of the 2022 Conference of the North American Chapter of the Association for Computational Linguistics: Human Language Technologies}}, \bibfield{editor}{\bibinfo{person}{Marine Carpuat}, \bibinfo{person}{Marie-Catherine de~Marneffe}, {and} \bibinfo{person}{Ivan~Vladimir Meza~Ruiz}} (Eds.). \bibinfo{publisher}{Association for Computational Linguistics}, \bibinfo{address}{Seattle, United States}, \bibinfo{pages}{2740--2750}.
\newblock
\urldef\tempurl%
\url{https://doi.org/10.18653/v1/2022.naacl-main.197}
\showDOI{\tempurl}


\bibitem[Kiritchenko and Mohammad(2017)]%
        {kiritchenko-mohammad-2017-best}
\bibfield{author}{\bibinfo{person}{Svetlana Kiritchenko} {and} \bibinfo{person}{Saif Mohammad}.} \bibinfo{year}{2017}\natexlab{}.
\newblock \showarticletitle{Best-Worst Scaling More Reliable than Rating Scales: A Case Study on Sentiment Intensity Annotation}. In \bibinfo{booktitle}{\emph{Proceedings of the 55th Annual Meeting of the Association for Computational Linguistics (Volume 2: Short Papers)}}, \bibfield{editor}{\bibinfo{person}{Regina Barzilay} {and} \bibinfo{person}{Min-Yen Kan}} (Eds.). \bibinfo{publisher}{Association for Computational Linguistics}, \bibinfo{address}{Vancouver, Canada}, \bibinfo{pages}{465--470}.
\newblock
\urldef\tempurl%
\url{https://doi.org/10.18653/v1/P17-2074}
\showDOI{\tempurl}


\bibitem[Kiritchenko and Mohammad(2016)]%
        {kiritchenko-mohammad-2016-capturing}
\bibfield{author}{\bibinfo{person}{Svetlana Kiritchenko} {and} \bibinfo{person}{Saif~M. Mohammad}.} \bibinfo{year}{2016}\natexlab{}.
\newblock \showarticletitle{Capturing Reliable Fine-Grained Sentiment Associations by Crowdsourcing and Best{--}Worst Scaling}. In \bibinfo{booktitle}{\emph{Proceedings of the 2016 Conference of the North {A}merican Chapter of the Association for Computational Linguistics: Human Language Technologies}}, \bibfield{editor}{\bibinfo{person}{Kevin Knight}, \bibinfo{person}{Ani Nenkova}, {and} \bibinfo{person}{Owen Rambow}} (Eds.). \bibinfo{publisher}{Association for Computational Linguistics}, \bibinfo{address}{San Diego, California}, \bibinfo{pages}{811--817}.
\newblock
\urldef\tempurl%
\url{https://doi.org/10.18653/v1/N16-1095}
\showDOI{\tempurl}


\bibitem[Kirk et~al\mbox{.}(2021)]%
        {kirk2021bias}
\bibfield{author}{\bibinfo{person}{Hannah Kirk}, \bibinfo{person}{Yennie Jun}, \bibinfo{person}{Haider Iqbal}, \bibinfo{person}{Elias Benussi}, \bibinfo{person}{Filippo Volpin}, \bibinfo{person}{Frederic~A. Dreyer}, \bibinfo{person}{Aleksandar Shtedritski}, {and} \bibinfo{person}{Yuki~M. Asano}.} \bibinfo{year}{2021}\natexlab{}.
\newblock \bibinfo{title}{Bias Out-of-the-Box: An Empirical Analysis of Intersectional Occupational Biases in Popular Generative Language Models}.
\newblock
\newblock
\showeprint[arxiv]{2102.04130}~[cs.CL]


\bibitem[Kirtane and Anand(2022)]%
        {kirtane-anand-2022-mitigating}
\bibfield{author}{\bibinfo{person}{Neeraja Kirtane} {and} \bibinfo{person}{Tanvi Anand}.} \bibinfo{year}{2022}\natexlab{}.
\newblock \showarticletitle{Mitigating Gender Stereotypes in {H}indi and {M}arathi}. In \bibinfo{booktitle}{\emph{Proceedings of the 4th Workshop on Gender Bias in Natural Language Processing (GeBNLP)}}, \bibfield{editor}{\bibinfo{person}{Christian Hardmeier}, \bibinfo{person}{Christine Basta}, \bibinfo{person}{Marta~R. Costa-juss{\`a}}, \bibinfo{person}{Gabriel Stanovsky}, {and} \bibinfo{person}{Hila Gonen}} (Eds.). \bibinfo{publisher}{Association for Computational Linguistics}, \bibinfo{address}{Seattle, Washington}, \bibinfo{pages}{145--150}.
\newblock
\urldef\tempurl%
\url{https://doi.org/10.18653/v1/2022.gebnlp-1.16}
\showDOI{\tempurl}


\bibitem[Kormilitzin et~al\mbox{.}(2023)]%
        {Kormilitzin2023API}
\bibfield{author}{\bibinfo{person}{Andrey Kormilitzin}, \bibinfo{person}{Nenad Tomasev}, \bibinfo{person}{Kevin~R. McKee}, {and} \bibinfo{person}{Dan~W. Joyce}.} \bibinfo{year}{2023}\natexlab{}.
\newblock \showarticletitle{A participatory initiative to include LGBT+ voices in AI for mental health}.
\newblock \bibinfo{journal}{\emph{Nature Medicine}}  \bibinfo{volume}{29} (\bibinfo{year}{2023}), \bibinfo{pages}{10--11}.
\newblock
\urldef\tempurl%
\url{https://api.semanticscholar.org/CorpusID:255748280}
\showURL{%
\tempurl}


\bibitem[Laskar et~al\mbox{.}(2023)]%
        {laskar-etal-2023-building}
\bibfield{author}{\bibinfo{person}{Md~Tahmid~Rahman Laskar}, \bibinfo{person}{Xue-Yong Fu}, \bibinfo{person}{Cheng Chen}, {and} \bibinfo{person}{Shashi Bhushan~TN}.} \bibinfo{year}{2023}\natexlab{}.
\newblock \showarticletitle{Building Real-World Meeting Summarization Systems using Large Language Models: A Practical Perspective}. In \bibinfo{booktitle}{\emph{Proceedings of the 2023 Conference on Empirical Methods in Natural Language Processing: Industry Track}}, \bibfield{editor}{\bibinfo{person}{Mingxuan Wang} {and} \bibinfo{person}{Imed Zitouni}} (Eds.). \bibinfo{publisher}{Association for Computational Linguistics}, \bibinfo{address}{Singapore}, \bibinfo{pages}{343--352}.
\newblock
\urldef\tempurl%
\url{https://doi.org/10.18653/v1/2023.emnlp-industry.33}
\showDOI{\tempurl}


\bibitem[Lauscher et~al\mbox{.}(2021)]%
        {lauscher-etal-2021-sustainable-modular}
\bibfield{author}{\bibinfo{person}{Anne Lauscher}, \bibinfo{person}{Tobias Lueken}, {and} \bibinfo{person}{Goran Glava{\v{s}}}.} \bibinfo{year}{2021}\natexlab{}.
\newblock \showarticletitle{Sustainable Modular Debiasing of Language Models}. In \bibinfo{booktitle}{\emph{Findings of the Association for Computational Linguistics: EMNLP 2021}}, \bibfield{editor}{\bibinfo{person}{Marie-Francine Moens}, \bibinfo{person}{Xuanjing Huang}, \bibinfo{person}{Lucia Specia}, {and} \bibinfo{person}{Scott Wen-tau Yih}} (Eds.). \bibinfo{publisher}{Association for Computational Linguistics}, \bibinfo{address}{Punta Cana, Dominican Republic}, \bibinfo{pages}{4782--4797}.
\newblock
\urldef\tempurl%
\url{https://doi.org/10.18653/v1/2021.findings-emnlp.411}
\showDOI{\tempurl}


\bibitem[Levy et~al\mbox{.}(2021)]%
        {levy-etal-2021-collecting-large}
\bibfield{author}{\bibinfo{person}{Shahar Levy}, \bibinfo{person}{Koren Lazar}, {and} \bibinfo{person}{Gabriel Stanovsky}.} \bibinfo{year}{2021}\natexlab{}.
\newblock \showarticletitle{Collecting a Large-Scale Gender Bias Dataset for Coreference Resolution and Machine Translation}. In \bibinfo{booktitle}{\emph{Findings of the Association for Computational Linguistics: EMNLP 2021}}, \bibfield{editor}{\bibinfo{person}{Marie-Francine Moens}, \bibinfo{person}{Xuanjing Huang}, \bibinfo{person}{Lucia Specia}, {and} \bibinfo{person}{Scott Wen-tau Yih}} (Eds.). \bibinfo{publisher}{Association for Computational Linguistics}, \bibinfo{address}{Punta Cana, Dominican Republic}, \bibinfo{pages}{2470--2480}.
\newblock
\urldef\tempurl%
\url{https://doi.org/10.18653/v1/2021.findings-emnlp.211}
\showDOI{\tempurl}


\bibitem[Lin et~al\mbox{.}(2019)]%
        {lin-etal-2019-choosing}
\bibfield{author}{\bibinfo{person}{Yu-Hsiang Lin}, \bibinfo{person}{Chian-Yu Chen}, \bibinfo{person}{Jean Lee}, \bibinfo{person}{Zirui Li}, \bibinfo{person}{Yuyan Zhang}, \bibinfo{person}{Mengzhou Xia}, \bibinfo{person}{Shruti Rijhwani}, \bibinfo{person}{Junxian He}, \bibinfo{person}{Zhisong Zhang}, \bibinfo{person}{Xuezhe Ma}, \bibinfo{person}{Antonios Anastasopoulos}, \bibinfo{person}{Patrick Littell}, {and} \bibinfo{person}{Graham Neubig}.} \bibinfo{year}{2019}\natexlab{}.
\newblock \showarticletitle{Choosing Transfer Languages for Cross-Lingual Learning}. In \bibinfo{booktitle}{\emph{Proceedings of the 57th Annual Meeting of the Association for Computational Linguistics}}, \bibfield{editor}{\bibinfo{person}{Anna Korhonen}, \bibinfo{person}{David Traum}, {and} \bibinfo{person}{Llu{\'\i}s M{\`a}rquez}} (Eds.). \bibinfo{publisher}{Association for Computational Linguistics}, \bibinfo{address}{Florence, Italy}, \bibinfo{pages}{3125--3135}.
\newblock
\urldef\tempurl%
\url{https://doi.org/10.18653/v1/P19-1301}
\showDOI{\tempurl}


\bibitem[Liu et~al\mbox{.}(2023)]%
        {liu-etal-2023-crossing}
\bibfield{author}{\bibinfo{person}{Emmy Liu}, \bibinfo{person}{Aditi Chaudhary}, {and} \bibinfo{person}{Graham Neubig}.} \bibinfo{year}{2023}\natexlab{}.
\newblock \showarticletitle{Crossing the Threshold: Idiomatic Machine Translation through Retrieval Augmentation and Loss Weighting}. In \bibinfo{booktitle}{\emph{Proceedings of the 2023 Conference on Empirical Methods in Natural Language Processing}}, \bibfield{editor}{\bibinfo{person}{Houda Bouamor}, \bibinfo{person}{Juan Pino}, {and} \bibinfo{person}{Kalika Bali}} (Eds.). \bibinfo{publisher}{Association for Computational Linguistics}, \bibinfo{address}{Singapore}, \bibinfo{pages}{15095--15111}.
\newblock
\urldef\tempurl%
\url{https://doi.org/10.18653/v1/2023.emnlp-main.933}
\showDOI{\tempurl}


\bibitem[Louviere(1991)]%
        {bws_jj}
\bibfield{author}{\bibinfo{person}{J.~J. Louviere}.} \bibinfo{year}{1991}\natexlab{}.
\newblock \bibinfo{title}{Best-worst scaling: A model for the largest difference judgments. Working Paper}.
\newblock
\newblock


\bibitem[Lucy et~al\mbox{.}(2023)]%
        {lucy2023onesizefitsall}
\bibfield{author}{\bibinfo{person}{Li Lucy}, \bibinfo{person}{Su~Lin Blodgett}, \bibinfo{person}{Milad Shokouhi}, \bibinfo{person}{Hanna Wallach}, {and} \bibinfo{person}{Alexandra Olteanu}.} \bibinfo{year}{2023}\natexlab{}.
\newblock \bibinfo{title}{"One-size-fits-all"? Observations and Expectations of NLG Systems Across Identity-Related Language Features}.
\newblock
\newblock
\showeprint[arxiv]{2310.15398}~[cs.CL]


\bibitem[M{'}hamdi et~al\mbox{.}(2023)]%
        {mhamdi-etal-2023-cross}
\bibfield{author}{\bibinfo{person}{Meryem M{'}hamdi}, \bibinfo{person}{Xiang Ren}, {and} \bibinfo{person}{Jonathan May}.} \bibinfo{year}{2023}\natexlab{}.
\newblock \showarticletitle{Cross-lingual Continual Learning}. In \bibinfo{booktitle}{\emph{Proceedings of the 61st Annual Meeting of the Association for Computational Linguistics (Volume 1: Long Papers)}}, \bibfield{editor}{\bibinfo{person}{Anna Rogers}, \bibinfo{person}{Jordan Boyd-Graber}, {and} \bibinfo{person}{Naoaki Okazaki}} (Eds.). \bibinfo{publisher}{Association for Computational Linguistics}, \bibinfo{address}{Toronto, Canada}, \bibinfo{pages}{3908--3943}.
\newblock
\urldef\tempurl%
\url{https://doi.org/10.18653/v1/2023.acl-long.217}
\showDOI{\tempurl}


\bibitem[Mieczkowski et~al\mbox{.}(2021)]%
        {10.1145/3449091}
\bibfield{author}{\bibinfo{person}{Hannah Mieczkowski}, \bibinfo{person}{Jeffrey~T. Hancock}, \bibinfo{person}{Mor Naaman}, \bibinfo{person}{Malte Jung}, {and} \bibinfo{person}{Jess Hohenstein}.} \bibinfo{year}{2021}\natexlab{}.
\newblock \showarticletitle{AI-Mediated Communication: Language Use and Interpersonal Effects in a Referential Communication Task}.
\newblock \bibinfo{journal}{\emph{Proc. ACM Hum.-Comput. Interact.}} \bibinfo{volume}{5}, \bibinfo{number}{CSCW1}, Article \bibinfo{articleno}{17} (\bibinfo{date}{apr} \bibinfo{year}{2021}), \bibinfo{numpages}{14}~pages.
\newblock
\urldef\tempurl%
\url{https://doi.org/10.1145/3449091}
\showDOI{\tempurl}


\bibitem[Mishra et~al\mbox{.}(2019)]%
        {pushkar-survey}
\bibfield{author}{\bibinfo{person}{Pushkar Mishra}, \bibinfo{person}{Helen Yannakoudakis}, {and} \bibinfo{person}{Ekaterina Shutova}.} \bibinfo{year}{2019}\natexlab{}.
\newblock \showarticletitle{Tackling Online Abuse: {A} Survey of Automated Abuse Detection Methods}.
\newblock \bibinfo{journal}{\emph{CoRR}}  \bibinfo{volume}{abs/1908.06024} (\bibinfo{year}{2019}).
\newblock
\showeprint[arXiv]{1908.06024}
\urldef\tempurl%
\url{http://arxiv.org/abs/1908.06024}
\showURL{%
\tempurl}


\bibitem[Mostafazadeh~Davani et~al\mbox{.}(2022)]%
        {davani-etal-2022-dealing}
\bibfield{author}{\bibinfo{person}{Aida Mostafazadeh~Davani}, \bibinfo{person}{Mark D{\'\i}az}, {and} \bibinfo{person}{Vinodkumar Prabhakaran}.} \bibinfo{year}{2022}\natexlab{}.
\newblock \showarticletitle{Dealing with Disagreements: Looking Beyond the Majority Vote in Subjective Annotations}.
\newblock \bibinfo{journal}{\emph{Transactions of the Association for Computational Linguistics}}  \bibinfo{volume}{10} (\bibinfo{year}{2022}), \bibinfo{pages}{92--110}.
\newblock
\urldef\tempurl%
\url{https://doi.org/10.1162/tacl_a_00449}
\showDOI{\tempurl}


\bibitem[Nadeem et~al\mbox{.}(2021)]%
        {nadeem-etal-2021-stereoset}
\bibfield{author}{\bibinfo{person}{Moin Nadeem}, \bibinfo{person}{Anna Bethke}, {and} \bibinfo{person}{Siva Reddy}.} \bibinfo{year}{2021}\natexlab{}.
\newblock \showarticletitle{{S}tereo{S}et: Measuring stereotypical bias in pretrained language models}. In \bibinfo{booktitle}{\emph{Proceedings of the 59th Annual Meeting of the Association for Computational Linguistics and the 11th International Joint Conference on Natural Language Processing (Volume 1: Long Papers)}}, \bibfield{editor}{\bibinfo{person}{Chengqing Zong}, \bibinfo{person}{Fei Xia}, \bibinfo{person}{Wenjie Li}, {and} \bibinfo{person}{Roberto Navigli}} (Eds.). \bibinfo{publisher}{Association for Computational Linguistics}, \bibinfo{address}{Online}, \bibinfo{pages}{5356--5371}.
\newblock
\urldef\tempurl%
\url{https://doi.org/10.18653/v1/2021.acl-long.416}
\showDOI{\tempurl}


\bibitem[Nangia et~al\mbox{.}(2020)]%
        {nangia-etal-2020-crows}
\bibfield{author}{\bibinfo{person}{Nikita Nangia}, \bibinfo{person}{Clara Vania}, \bibinfo{person}{Rasika Bhalerao}, {and} \bibinfo{person}{Samuel~R. Bowman}.} \bibinfo{year}{2020}\natexlab{}.
\newblock \showarticletitle{{C}row{S}-Pairs: A Challenge Dataset for Measuring Social Biases in Masked Language Models}. In \bibinfo{booktitle}{\emph{Proceedings of the 2020 Conference on Empirical Methods in Natural Language Processing (EMNLP)}}, \bibfield{editor}{\bibinfo{person}{Bonnie Webber}, \bibinfo{person}{Trevor Cohn}, \bibinfo{person}{Yulan He}, {and} \bibinfo{person}{Yang Liu}} (Eds.). \bibinfo{publisher}{Association for Computational Linguistics}, \bibinfo{address}{Online}, \bibinfo{pages}{1953--1967}.
\newblock
\urldef\tempurl%
\url{https://doi.org/10.18653/v1/2020.emnlp-main.154}
\showDOI{\tempurl}


\bibitem[Ntoutsi et~al\mbox{.}(2020)]%
        {ntoutsi2020bias}
\bibfield{author}{\bibinfo{person}{Eirini Ntoutsi}, \bibinfo{person}{Pavlos Fafalios}, \bibinfo{person}{Ujwal Gadiraju}, \bibinfo{person}{Vasileios Iosifidis}, \bibinfo{person}{Wolfgang Nejdl}, \bibinfo{person}{Maria-Esther Vidal}, \bibinfo{person}{Salvatore Ruggieri}, \bibinfo{person}{Franco Turini}, \bibinfo{person}{Symeon Papadopoulos}, \bibinfo{person}{Emmanouil Krasanakis}, {et~al\mbox{.}}} \bibinfo{year}{2020}\natexlab{}.
\newblock \showarticletitle{Bias in data-driven artificial intelligence systems—An introductory survey}.
\newblock \bibinfo{journal}{\emph{Wiley Interdisciplinary Reviews: Data Mining and Knowledge Discovery}} \bibinfo{volume}{10}, \bibinfo{number}{3} (\bibinfo{year}{2020}), \bibinfo{pages}{e1356}.
\newblock


\bibitem[Orgad and Belinkov(2022)]%
        {orgad-belinkov-2022-choose}
\bibfield{author}{\bibinfo{person}{Hadas Orgad} {and} \bibinfo{person}{Yonatan Belinkov}.} \bibinfo{year}{2022}\natexlab{}.
\newblock \showarticletitle{Choose Your Lenses: Flaws in Gender Bias Evaluation}. In \bibinfo{booktitle}{\emph{Proceedings of the 4th Workshop on Gender Bias in Natural Language Processing (GeBNLP)}}, \bibfield{editor}{\bibinfo{person}{Christian Hardmeier}, \bibinfo{person}{Christine Basta}, \bibinfo{person}{Marta~R. Costa-juss{\`a}}, \bibinfo{person}{Gabriel Stanovsky}, {and} \bibinfo{person}{Hila Gonen}} (Eds.). \bibinfo{publisher}{Association for Computational Linguistics}, \bibinfo{address}{Seattle, Washington}, \bibinfo{pages}{151--167}.
\newblock
\urldef\tempurl%
\url{https://doi.org/10.18653/v1/2022.gebnlp-1.17}
\showDOI{\tempurl}


\bibitem[Orme(2009)]%
        {omre}
\bibfield{author}{\bibinfo{person}{B. Orme}.} \bibinfo{year}{2009}\natexlab{}.
\newblock \bibinfo{title}{Maxdiff analysis: Simple counting,individual-level logit, and HB. Sawtooth Software, Inc}.
\newblock
\newblock


\bibitem[Pei and Jurgens(2020)]%
        {pei-jurgens-2020-quantifying}
\bibfield{author}{\bibinfo{person}{Jiaxin Pei} {and} \bibinfo{person}{David Jurgens}.} \bibinfo{year}{2020}\natexlab{}.
\newblock \showarticletitle{Quantifying Intimacy in Language}. In \bibinfo{booktitle}{\emph{Proceedings of the 2020 Conference on Empirical Methods in Natural Language Processing (EMNLP)}}, \bibfield{editor}{\bibinfo{person}{Bonnie Webber}, \bibinfo{person}{Trevor Cohn}, \bibinfo{person}{Yulan He}, {and} \bibinfo{person}{Yang Liu}} (Eds.). \bibinfo{publisher}{Association for Computational Linguistics}, \bibinfo{address}{Online}, \bibinfo{pages}{5307--5326}.
\newblock
\urldef\tempurl%
\url{https://doi.org/10.18653/v1/2020.emnlp-main.428}
\showDOI{\tempurl}


\bibitem[Philippy et~al\mbox{.}(2023)]%
        {philippy-etal-2023-towards}
\bibfield{author}{\bibinfo{person}{Fred Philippy}, \bibinfo{person}{Siwen Guo}, {and} \bibinfo{person}{Shohreh Haddadan}.} \bibinfo{year}{2023}\natexlab{}.
\newblock \showarticletitle{Towards a Common Understanding of Contributing Factors for Cross-Lingual Transfer in Multilingual Language Models: A Review}. In \bibinfo{booktitle}{\emph{Proceedings of the 61st Annual Meeting of the Association for Computational Linguistics (Volume 1: Long Papers)}}, \bibfield{editor}{\bibinfo{person}{Anna Rogers}, \bibinfo{person}{Jordan Boyd-Graber}, {and} \bibinfo{person}{Naoaki Okazaki}} (Eds.). \bibinfo{publisher}{Association for Computational Linguistics}, \bibinfo{address}{Toronto, Canada}, \bibinfo{pages}{5877--5891}.
\newblock
\urldef\tempurl%
\url{https://doi.org/10.18653/v1/2023.acl-long.323}
\showDOI{\tempurl}


\bibitem[Prabhakaran et~al\mbox{.}(2022)]%
        {prabhakaran2022cultural}
\bibfield{author}{\bibinfo{person}{Vinodkumar Prabhakaran}, \bibinfo{person}{Rida Qadri}, {and} \bibinfo{person}{Ben Hutchinson}.} \bibinfo{year}{2022}\natexlab{}.
\newblock \bibinfo{title}{Cultural Incongruencies in Artificial Intelligence}.
\newblock
\newblock
\showeprint[arxiv]{2211.13069}~[cs.CY]


\bibitem[Pratham(2022)]%
        {ASER2022}
\bibfield{author}{\bibinfo{person}{Pratham}.} \bibinfo{year}{2022}\natexlab{}.
\newblock \bibinfo{title}{Annual Status of Education Report 2022}.
\newblock
\newblock
\urldef\tempurl%
\url{https://asercentre.org/aser-2022/}
\showURL{%
\tempurl}
\newblock
\shownote{Accessed on 09/13/2023}.


\bibitem[Pujari et~al\mbox{.}(2020)]%
        {10.1145/3377713.3377792}
\bibfield{author}{\bibinfo{person}{Arun~K. Pujari}, \bibinfo{person}{Ansh Mittal}, \bibinfo{person}{Anshuman Padhi}, \bibinfo{person}{Anshul Jain}, \bibinfo{person}{Mukesh Jadon}, {and} \bibinfo{person}{Vikas Kumar}.} \bibinfo{year}{2020}\natexlab{}.
\newblock \showarticletitle{Debiasing Gender biased Hindi Words with Word-embedding}. In \bibinfo{booktitle}{\emph{Proceedings of the 2019 2nd International Conference on Algorithms, Computing and Artificial Intelligence}} (Sanya, China) \emph{(\bibinfo{series}{ACAI '19})}. \bibinfo{publisher}{Association for Computing Machinery}, \bibinfo{address}{New York, NY, USA}, \bibinfo{pages}{450–456}.
\newblock
\showISBNx{9781450372619}
\urldef\tempurl%
\url{https://doi.org/10.1145/3377713.3377792}
\showDOI{\tempurl}


\bibitem[Queerinai et~al\mbox{.}(2023)]%
        {10.1145/3593013.3594134}
\bibfield{author}{\bibinfo{person}{Organizers~Of Queerinai}, \bibinfo{person}{Anaelia Ovalle}, \bibinfo{person}{Arjun Subramonian}, \bibinfo{person}{Ashwin Singh}, \bibinfo{person}{Claas Voelcker}, \bibinfo{person}{Danica~J. Sutherland}, \bibinfo{person}{Davide Locatelli}, \bibinfo{person}{Eva Breznik}, \bibinfo{person}{Filip Klubicka}, \bibinfo{person}{Hang Yuan}, \bibinfo{person}{Hetvi J}, \bibinfo{person}{Huan Zhang}, \bibinfo{person}{Jaidev Shriram}, \bibinfo{person}{Kruno Lehman}, \bibinfo{person}{Luca Soldaini}, \bibinfo{person}{Maarten Sap}, \bibinfo{person}{Marc~Peter Deisenroth}, \bibinfo{person}{Maria~Leonor Pacheco}, \bibinfo{person}{Maria Ryskina}, \bibinfo{person}{Martin Mundt}, \bibinfo{person}{Milind Agarwal}, \bibinfo{person}{Nyx Mclean}, \bibinfo{person}{Pan Xu}, \bibinfo{person}{A Pranav}, \bibinfo{person}{Raj Korpan}, \bibinfo{person}{Ruchira Ray}, \bibinfo{person}{Sarah Mathew}, \bibinfo{person}{Sarthak Arora}, \bibinfo{person}{St John}, \bibinfo{person}{Tanvi Anand},
  \bibinfo{person}{Vishakha Agrawal}, \bibinfo{person}{William Agnew}, \bibinfo{person}{Yanan Long}, \bibinfo{person}{Zijie~J. Wang}, \bibinfo{person}{Zeerak Talat}, \bibinfo{person}{Avijit Ghosh}, \bibinfo{person}{Nathaniel Dennler}, \bibinfo{person}{Michael Noseworthy}, \bibinfo{person}{Sharvani Jha}, \bibinfo{person}{Emi Baylor}, \bibinfo{person}{Aditya Joshi}, \bibinfo{person}{Natalia~Y. Bilenko}, \bibinfo{person}{Andrew Mcnamara}, \bibinfo{person}{Raphael Gontijo-Lopes}, \bibinfo{person}{Alex Markham}, \bibinfo{person}{Evyn Dong}, \bibinfo{person}{Jackie Kay}, \bibinfo{person}{Manu Saraswat}, \bibinfo{person}{Nikhil Vytla}, {and} \bibinfo{person}{Luke Stark}.} \bibinfo{year}{2023}\natexlab{}.
\newblock \showarticletitle{Queer In AI: A Case Study in Community-Led Participatory AI}. In \bibinfo{booktitle}{\emph{Proceedings of the 2023 ACM Conference on Fairness, Accountability, and Transparency}} (Chicago, IL, USA) \emph{(\bibinfo{series}{FAccT '23})}. \bibinfo{publisher}{Association for Computing Machinery}, \bibinfo{address}{New York, NY, USA}, \bibinfo{pages}{1882–1895}.
\newblock
\showISBNx{9798400701924}
\urldef\tempurl%
\url{https://doi.org/10.1145/3593013.3594134}
\showDOI{\tempurl}


\bibitem[Ramesh et~al\mbox{.}(2021)]%
        {ramesh-etal-2021-evaluating}
\bibfield{author}{\bibinfo{person}{Krithika Ramesh}, \bibinfo{person}{Gauri Gupta}, {and} \bibinfo{person}{Sanjay Singh}.} \bibinfo{year}{2021}\natexlab{}.
\newblock \showarticletitle{Evaluating Gender Bias in {H}indi-{E}nglish Machine Translation}. In \bibinfo{booktitle}{\emph{Proceedings of the 3rd Workshop on Gender Bias in Natural Language Processing}}, \bibfield{editor}{\bibinfo{person}{Marta Costa-jussa}, \bibinfo{person}{Hila Gonen}, \bibinfo{person}{Christian Hardmeier}, {and} \bibinfo{person}{Kellie Webster}} (Eds.). \bibinfo{publisher}{Association for Computational Linguistics}, \bibinfo{address}{Online}, \bibinfo{pages}{16--23}.
\newblock
\urldef\tempurl%
\url{https://doi.org/10.18653/v1/2021.gebnlp-1.3}
\showDOI{\tempurl}


\bibitem[Ramesh et~al\mbox{.}(2023)]%
        {ramesh-etal-2023-fairness}
\bibfield{author}{\bibinfo{person}{Krithika Ramesh}, \bibinfo{person}{Sunayana Sitaram}, {and} \bibinfo{person}{Monojit Choudhury}.} \bibinfo{year}{2023}\natexlab{}.
\newblock \showarticletitle{Fairness in Language Models Beyond {E}nglish: Gaps and Challenges}. In \bibinfo{booktitle}{\emph{Findings of the Association for Computational Linguistics: EACL 2023}}, \bibfield{editor}{\bibinfo{person}{Andreas Vlachos} {and} \bibinfo{person}{Isabelle Augenstein}} (Eds.). \bibinfo{publisher}{Association for Computational Linguistics}, \bibinfo{address}{Dubrovnik, Croatia}, \bibinfo{pages}{2106--2119}.
\newblock
\urldef\tempurl%
\url{https://doi.org/10.18653/v1/2023.findings-eacl.157}
\showDOI{\tempurl}


\bibitem[Raza et~al\mbox{.}(2024)]%
        {DBLP:journals/eswa/RazaGRBD24}
\bibfield{author}{\bibinfo{person}{Shaina Raza}, \bibinfo{person}{Muskan Garg}, \bibinfo{person}{Deepak~John Reji}, \bibinfo{person}{Syed~Raza Bashir}, {and} \bibinfo{person}{Chen Ding}.} \bibinfo{year}{2024}\natexlab{}.
\newblock \showarticletitle{Nbias: {A} natural language processing framework for {BIAS} identification in text}.
\newblock \bibinfo{journal}{\emph{Expert Syst. Appl.}} \bibinfo{volume}{237}, \bibinfo{number}{Part {B}} (\bibinfo{year}{2024}), \bibinfo{pages}{121542}.
\newblock
\urldef\tempurl%
\url{https://doi.org/10.1016/J.ESWA.2023.121542}
\showDOI{\tempurl}


\bibitem[Rudinger et~al\mbox{.}(2018)]%
        {rudinger-etal-2018-gender}
\bibfield{author}{\bibinfo{person}{Rachel Rudinger}, \bibinfo{person}{Jason Naradowsky}, \bibinfo{person}{Brian Leonard}, {and} \bibinfo{person}{Benjamin Van~Durme}.} \bibinfo{year}{2018}\natexlab{}.
\newblock \showarticletitle{Gender Bias in Coreference Resolution}. In \bibinfo{booktitle}{\emph{Proceedings of the 2018 Conference of the North {A}merican Chapter of the Association for Computational Linguistics: Human Language Technologies, Volume 2 (Short Papers)}}, \bibfield{editor}{\bibinfo{person}{Marilyn Walker}, \bibinfo{person}{Heng Ji}, {and} \bibinfo{person}{Amanda Stent}} (Eds.). \bibinfo{publisher}{Association for Computational Linguistics}, \bibinfo{address}{New Orleans, Louisiana}, \bibinfo{pages}{8--14}.
\newblock
\urldef\tempurl%
\url{https://doi.org/10.18653/v1/N18-2002}
\showDOI{\tempurl}


\bibitem[Sharifi~Noorian et~al\mbox{.}(2023)]%
        {sharifi2023perspective}
\bibfield{author}{\bibinfo{person}{Shahin Sharifi~Noorian}, \bibinfo{person}{Sihang Qiu}, \bibinfo{person}{Burcu Sayin}, \bibinfo{person}{Agathe Balayn}, \bibinfo{person}{Ujwal Gadiraju}, \bibinfo{person}{Jie Yang}, {and} \bibinfo{person}{Alessandro Bozzon}.} \bibinfo{year}{2023}\natexlab{}.
\newblock \showarticletitle{Perspective: leveraging human understanding for identifying and characterizing image atypicality}. In \bibinfo{booktitle}{\emph{Proceedings of the 28th International Conference on Intelligent User Interfaces}}. \bibinfo{pages}{650--663}.
\newblock


\bibitem[Song et~al\mbox{.}(2023)]%
        {song2023restgpt}
\bibfield{author}{\bibinfo{person}{Yifan Song}, \bibinfo{person}{Weimin Xiong}, \bibinfo{person}{Dawei Zhu}, \bibinfo{person}{Wenhao Wu}, \bibinfo{person}{Han Qian}, \bibinfo{person}{Mingbo Song}, \bibinfo{person}{Hailiang Huang}, \bibinfo{person}{Cheng Li}, \bibinfo{person}{Ke Wang}, \bibinfo{person}{Rong Yao}, \bibinfo{person}{Ye Tian}, {and} \bibinfo{person}{Sujian Li}.} \bibinfo{year}{2023}\natexlab{}.
\newblock \showarticletitle{RestGPT: Connecting Large Language Models with Real-World RESTful APIs}.
\newblock \bibinfo{journal}{\emph{arXiv preprint arXiv: 2306.06624}} (\bibinfo{year}{2023}).
\newblock


\bibitem[Spinde et~al\mbox{.}(2021)]%
        {spinde-etal-2021-neural-media}
\bibfield{author}{\bibinfo{person}{Timo Spinde}, \bibinfo{person}{Manuel Plank}, \bibinfo{person}{Jan-David Krieger}, \bibinfo{person}{Terry Ruas}, \bibinfo{person}{Bela Gipp}, {and} \bibinfo{person}{Akiko Aizawa}.} \bibinfo{year}{2021}\natexlab{}.
\newblock \showarticletitle{Neural Media Bias Detection Using Distant Supervision With {BABE} - Bias Annotations By Experts}. In \bibinfo{booktitle}{\emph{Findings of the Association for Computational Linguistics: EMNLP 2021}}, \bibfield{editor}{\bibinfo{person}{Marie-Francine Moens}, \bibinfo{person}{Xuanjing Huang}, \bibinfo{person}{Lucia Specia}, {and} \bibinfo{person}{Scott Wen-tau Yih}} (Eds.). \bibinfo{publisher}{Association for Computational Linguistics}, \bibinfo{address}{Punta Cana, Dominican Republic}, \bibinfo{pages}{1166--1177}.
\newblock
\urldef\tempurl%
\url{https://doi.org/10.18653/v1/2021.findings-emnlp.101}
\showDOI{\tempurl}


\bibitem[Stanczak and Augenstein(2021)]%
        {stanczak2021survey}
\bibfield{author}{\bibinfo{person}{Karolina Stanczak} {and} \bibinfo{person}{Isabelle Augenstein}.} \bibinfo{year}{2021}\natexlab{}.
\newblock \showarticletitle{A Survey on Gender Bias in Natural Language Processing}.
\newblock \bibinfo{journal}{\emph{arXiv preprint arXiv: 2112.14168}} (\bibinfo{year}{2021}).
\newblock


\bibitem[Stanovsky et~al\mbox{.}(2019)]%
        {stanovsky-etal-2019-evaluating}
\bibfield{author}{\bibinfo{person}{Gabriel Stanovsky}, \bibinfo{person}{Noah~A. Smith}, {and} \bibinfo{person}{Luke Zettlemoyer}.} \bibinfo{year}{2019}\natexlab{}.
\newblock \showarticletitle{Evaluating Gender Bias in Machine Translation}. In \bibinfo{booktitle}{\emph{Proceedings of the 57th Annual Meeting of the Association for Computational Linguistics}}, \bibfield{editor}{\bibinfo{person}{Anna Korhonen}, \bibinfo{person}{David Traum}, {and} \bibinfo{person}{Llu{\'\i}s M{\`a}rquez}} (Eds.). \bibinfo{publisher}{Association for Computational Linguistics}, \bibinfo{address}{Florence, Italy}, \bibinfo{pages}{1679--1684}.
\newblock
\urldef\tempurl%
\url{https://doi.org/10.18653/v1/P19-1164}
\showDOI{\tempurl}


\bibitem[Statista(2022)]%
        {india-social}
\bibfield{author}{\bibinfo{person}{Statista}.} \bibinfo{year}{2022}\natexlab{}.
\newblock \bibinfo{title}{Use of social media platforms among people in India as of January 2022, by locality}.
\newblock
\newblock
\urldef\tempurl%
\url{https://www.statista.com/statistics/1388563/india-social-media-platform-usage-by-locality/}
\showURL{%
\tempurl}
\newblock
\shownote{Accessed: 2024-01-02}.


\bibitem[Statista(2023)]%
        {reddit-demo}
\bibfield{author}{\bibinfo{person}{Statista}.} \bibinfo{year}{2023}\natexlab{}.
\newblock \bibinfo{title}{Regional distribution of desktop traffic to Reddit.com as of April 2023 by country}.
\newblock
\newblock
\urldef\tempurl%
\url{https://www.statista.com/statistics/325144/reddit-global-active-user-distribution/}
\showURL{%
\tempurl}
\newblock
\shownote{Accessed: 2024-01-02}.


\bibitem[Suresh et~al\mbox{.}(2022)]%
        {10.1145/3531146.3533132}
\bibfield{author}{\bibinfo{person}{Harini Suresh}, \bibinfo{person}{Rajiv Movva}, \bibinfo{person}{Amelia~Lee Dogan}, \bibinfo{person}{Rahul Bhargava}, \bibinfo{person}{Isadora Cruxen}, \bibinfo{person}{Angeles~Martinez Cuba}, \bibinfo{person}{Guilia Taurino}, \bibinfo{person}{Wonyoung So}, {and} \bibinfo{person}{Catherine D'Ignazio}.} \bibinfo{year}{2022}\natexlab{}.
\newblock \showarticletitle{Towards Intersectional Feminist and Participatory ML: A Case Study in Supporting Feminicide Counterdata Collection}. In \bibinfo{booktitle}{\emph{Proceedings of the 2022 ACM Conference on Fairness, Accountability, and Transparency}} (Seoul, Republic of Korea) \emph{(\bibinfo{series}{FAccT '22})}. \bibinfo{publisher}{Association for Computing Machinery}, \bibinfo{address}{New York, NY, USA}, \bibinfo{pages}{667–678}.
\newblock
\showISBNx{9781450393522}
\urldef\tempurl%
\url{https://doi.org/10.1145/3531146.3533132}
\showDOI{\tempurl}


\bibitem[Trajanovski et~al\mbox{.}(2021)]%
        {trajanovski-etal-2021-text}
\bibfield{author}{\bibinfo{person}{Stojan Trajanovski}, \bibinfo{person}{Chad Atalla}, \bibinfo{person}{Kunho Kim}, \bibinfo{person}{Vipul Agarwal}, \bibinfo{person}{Milad Shokouhi}, {and} \bibinfo{person}{Chris Quirk}.} \bibinfo{year}{2021}\natexlab{}.
\newblock \showarticletitle{When does text prediction benefit from additional context? An exploration of contextual signals for chat and email messages}. In \bibinfo{booktitle}{\emph{Proceedings of the 2021 Conference of the North American Chapter of the Association for Computational Linguistics: Human Language Technologies: Industry Papers}}, \bibfield{editor}{\bibinfo{person}{Young-bum Kim}, \bibinfo{person}{Yunyao Li}, {and} \bibinfo{person}{Owen Rambow}} (Eds.). \bibinfo{publisher}{Association for Computational Linguistics}, \bibinfo{address}{Online}, \bibinfo{pages}{1--9}.
\newblock
\urldef\tempurl%
\url{https://doi.org/10.18653/v1/2021.naacl-industry.1}
\showDOI{\tempurl}


\bibitem[Vanmassenhove et~al\mbox{.}(2021)]%
        {vanmassenhove-etal-2021-machine}
\bibfield{author}{\bibinfo{person}{Eva Vanmassenhove}, \bibinfo{person}{Dimitar Shterionov}, {and} \bibinfo{person}{Matthew Gwilliam}.} \bibinfo{year}{2021}\natexlab{}.
\newblock \showarticletitle{Machine Translationese: Effects of Algorithmic Bias on Linguistic Complexity in Machine Translation}. In \bibinfo{booktitle}{\emph{Proceedings of the 16th Conference of the European Chapter of the Association for Computational Linguistics: Main Volume}}, \bibfield{editor}{\bibinfo{person}{Paola Merlo}, \bibinfo{person}{Jorg Tiedemann}, {and} \bibinfo{person}{Reut Tsarfaty}} (Eds.). \bibinfo{publisher}{Association for Computational Linguistics}, \bibinfo{address}{Online}, \bibinfo{pages}{2203--2213}.
\newblock
\urldef\tempurl%
\url{https://doi.org/10.18653/v1/2021.eacl-main.188}
\showDOI{\tempurl}


\bibitem[Varghese(2022)]%
        {india-eng}
\bibfield{author}{\bibinfo{person}{Ajit Varghese}.} \bibinfo{year}{2022}\natexlab{}.
\newblock \bibinfo{title}{Celebrating Bharat’s digital journey@75: The rapid increase in language-first users on social media}.
\newblock
\newblock
\urldef\tempurl%
\url{https://timesofindia.indiatimes.com/blogs/voices/celebrating-bharats-digital-journey75-the-rapid-increase-in-language-first-users-on-social-media/}
\showURL{%
\tempurl}
\newblock
\shownote{Accessed: 2024-01-02}.


\bibitem[Vashistha et~al\mbox{.}(2017)]%
        {vashistha_respeak}
\bibfield{author}{\bibinfo{person}{Aditya Vashistha}, \bibinfo{person}{Pooja Sethi}, {and} \bibinfo{person}{Richard Anderson}.} \bibinfo{year}{2017}\natexlab{}.
\newblock \showarticletitle{Respeak: A Voice-Based, Crowd-Powered Speech Transcription System}. In \bibinfo{booktitle}{\emph{Proceedings of the 2017 CHI Conference on Human Factors in Computing Systems}} (Denver, Colorado, USA) \emph{(\bibinfo{series}{CHI '17})}. \bibinfo{publisher}{Association for Computing Machinery}, \bibinfo{address}{New York, NY, USA}, \bibinfo{pages}{1855–1866}.
\newblock
\showISBNx{9781450346559}
\urldef\tempurl%
\url{https://doi.org/10.1145/3025453.3025640}
\showDOI{\tempurl}


\bibitem[von Ahn et~al\mbox{.}(2006)]%
        {10.1145/1124772.1124784}
\bibfield{author}{\bibinfo{person}{Luis von Ahn}, \bibinfo{person}{Mihir Kedia}, {and} \bibinfo{person}{Manuel Blum}.} \bibinfo{year}{2006}\natexlab{}.
\newblock \showarticletitle{Verbosity: A Game for Collecting Common-Sense Facts}. In \bibinfo{booktitle}{\emph{Proceedings of the SIGCHI Conference on Human Factors in Computing Systems}} (Montr\'{e}al, Qu\'{e}bec, Canada) \emph{(\bibinfo{series}{CHI '06})}. \bibinfo{publisher}{Association for Computing Machinery}, \bibinfo{address}{New York, NY, USA}, \bibinfo{pages}{75–78}.
\newblock
\showISBNx{1595933727}
\urldef\tempurl%
\url{https://doi.org/10.1145/1124772.1124784}
\showDOI{\tempurl}


\bibitem[Waseem and Hovy(2016)]%
        {waseem-hovy-2016-hateful}
\bibfield{author}{\bibinfo{person}{Zeerak Waseem} {and} \bibinfo{person}{Dirk Hovy}.} \bibinfo{year}{2016}\natexlab{}.
\newblock \showarticletitle{Hateful Symbols or Hateful People? Predictive Features for Hate Speech Detection on {T}witter}. In \bibinfo{booktitle}{\emph{Proceedings of the {NAACL} Student Research Workshop}}, \bibfield{editor}{\bibinfo{person}{Jacob Andreas}, \bibinfo{person}{Eunsol Choi}, {and} \bibinfo{person}{Angeliki Lazaridou}} (Eds.). \bibinfo{publisher}{Association for Computational Linguistics}, \bibinfo{address}{San Diego, California}, \bibinfo{pages}{88--93}.
\newblock
\urldef\tempurl%
\url{https://doi.org/10.18653/v1/N16-2013}
\showDOI{\tempurl}


\bibitem[Webster et~al\mbox{.}(2020)]%
        {50755}
\bibfield{author}{\bibinfo{person}{Kellie Webster}, \bibinfo{person}{Xuezhi Wang}, \bibinfo{person}{Ian Tenney}, \bibinfo{person}{Alex Beutel}, \bibinfo{person}{Emily Pitler}, \bibinfo{person}{Ellie Pavlick}, \bibinfo{person}{Jilin Chen}, \bibinfo{person}{Ed~H. Chi}, {and} \bibinfo{person}{Slav Petrov}.} \bibinfo{year}{2020}\natexlab{}.
\newblock \bibinfo{booktitle}{\emph{Measuring and Reducing Gendered Correlations in Pre-trained Models}}.
\newblock \bibinfo{type}{{T}echnical {R}eport}.
\newblock
\urldef\tempurl%
\url{https://arxiv.org/abs/2010.06032}
\showURL{%
\tempurl}


\bibitem[Zhang et~al\mbox{.}(2023)]%
        {zhang2023corgipm}
\bibfield{author}{\bibinfo{person}{Ge Zhang}, \bibinfo{person}{Yizhi Li}, \bibinfo{person}{Yaoyao Wu}, \bibinfo{person}{Linyuan Zhang}, \bibinfo{person}{Chenghua Lin}, \bibinfo{person}{Jiayi Geng}, \bibinfo{person}{Shi Wang}, {and} \bibinfo{person}{Jie Fu}.} \bibinfo{year}{2023}\natexlab{}.
\newblock \bibinfo{title}{CORGI-PM: A Chinese Corpus For Gender Bias Probing and Mitigation}.
\newblock
\newblock
\showeprint[arxiv]{2301.00395}~[cs.CL]


\bibitem[Zhang and Toral(2019)]%
        {zhang-toral-2019-effect}
\bibfield{author}{\bibinfo{person}{Mike Zhang} {and} \bibinfo{person}{Antonio Toral}.} \bibinfo{year}{2019}\natexlab{}.
\newblock \showarticletitle{The Effect of Translationese in Machine Translation Test Sets}. In \bibinfo{booktitle}{\emph{Proceedings of the Fourth Conference on Machine Translation (Volume 1: Research Papers)}}, \bibfield{editor}{\bibinfo{person}{Ond{\v{r}}ej Bojar}, \bibinfo{person}{Rajen Chatterjee}, \bibinfo{person}{Christian Federmann}, \bibinfo{person}{Mark Fishel}, \bibinfo{person}{Yvette Graham}, \bibinfo{person}{Barry Haddow}, \bibinfo{person}{Matthias Huck}, \bibinfo{person}{Antonio~Jimeno Yepes}, \bibinfo{person}{Philipp Koehn}, \bibinfo{person}{Andr{\'e} Martins}, \bibinfo{person}{Christof Monz}, \bibinfo{person}{Matteo Negri}, \bibinfo{person}{Aur{\'e}lie N{\'e}v{\'e}ol}, \bibinfo{person}{Mariana Neves}, \bibinfo{person}{Matt Post}, \bibinfo{person}{Marco Turchi}, {and} \bibinfo{person}{Karin Verspoor}} (Eds.). \bibinfo{publisher}{Association for Computational Linguistics}, \bibinfo{address}{Florence, Italy}, \bibinfo{pages}{73--81}.
\newblock
\urldef\tempurl%
\url{https://doi.org/10.18653/v1/W19-5208}
\showDOI{\tempurl}


\bibitem[Zhao et~al\mbox{.}(2018)]%
        {zhao-etal-2018-gender}
\bibfield{author}{\bibinfo{person}{Jieyu Zhao}, \bibinfo{person}{Tianlu Wang}, \bibinfo{person}{Mark Yatskar}, \bibinfo{person}{Vicente Ordonez}, {and} \bibinfo{person}{Kai-Wei Chang}.} \bibinfo{year}{2018}\natexlab{}.
\newblock \showarticletitle{Gender Bias in Coreference Resolution: Evaluation and Debiasing Methods}. In \bibinfo{booktitle}{\emph{Proceedings of the 2018 Conference of the North {A}merican Chapter of the Association for Computational Linguistics: Human Language Technologies, Volume 2 (Short Papers)}}, \bibfield{editor}{\bibinfo{person}{Marilyn Walker}, \bibinfo{person}{Heng Ji}, {and} \bibinfo{person}{Amanda Stent}} (Eds.). \bibinfo{publisher}{Association for Computational Linguistics}, \bibinfo{address}{New Orleans, Louisiana}, \bibinfo{pages}{15--20}.
\newblock
\urldef\tempurl%
\url{https://doi.org/10.18653/v1/N18-2003}
\showDOI{\tempurl}


\bibitem[Zhou et~al\mbox{.}(2021)]%
        {zhou-etal-2021-challenges}
\bibfield{author}{\bibinfo{person}{Xuhui Zhou}, \bibinfo{person}{Maarten Sap}, \bibinfo{person}{Swabha Swayamdipta}, \bibinfo{person}{Yejin Choi}, {and} \bibinfo{person}{Noah Smith}.} \bibinfo{year}{2021}\natexlab{}.
\newblock \showarticletitle{Challenges in Automated Debiasing for Toxic Language Detection}. In \bibinfo{booktitle}{\emph{Proceedings of the 16th Conference of the European Chapter of the Association for Computational Linguistics: Main Volume}}, \bibfield{editor}{\bibinfo{person}{Paola Merlo}, \bibinfo{person}{Jorg Tiedemann}, {and} \bibinfo{person}{Reut Tsarfaty}} (Eds.). \bibinfo{publisher}{Association for Computational Linguistics}, \bibinfo{address}{Online}, \bibinfo{pages}{3143--3155}.
\newblock
\urldef\tempurl%
\url{https://doi.org/10.18653/v1/2021.eacl-main.274}
\showDOI{\tempurl}


\end{thebibliography}
\end{document}